\documentclass[11pt]{article}

\usepackage[preprint]{acl}

\usepackage{times}
\usepackage{latexsym}
\usepackage[T1]{fontenc}
\usepackage[utf8]{inputenc}
\usepackage{microtype}
\usepackage{inconsolata}
\usepackage{graphicx}
\usepackage{booktabs}
\usepackage{multirow}
\usepackage{amsmath}
\usepackage{amssymb}

\usepackage{fvextra}
\RecustomVerbatimEnvironment{verbatim}{Verbatim}{breaklines=true,breakanywhere=true}

\title{DynamicMCPBench: A Trace-Grounded, Effect-Scored Benchmark for LLM Agents over Live MCP Servers}

\author{
  Jerzy Kami\'nski$^{1,\dagger,*}$ \quad Ilya Galyukshev$^{1,2,\dagger}$ \quad Artem Kuznetsov$^{1}$ \quad Sergey Chuprin$^{1}$ \\[2pt]
  Kirill Redko$^{1}$ \quad Aidar Shumbalov$^{1}$ \quad Anna Kalyuzhnaya$^{1}$ \\[4pt]
  $^{1}$ITMO University \quad $^{2}$Central University \\[4pt]
  $^{\dagger}$Equal contribution. \quad $^{*}$Correspondence: \texttt{jkaminski@niuitmo.ru}
}

\begin{document}
\maketitle

\begin{abstract}
Large language model (LLM) agents are increasingly deployed over Model
Context Protocol (MCP) servers, yet the benchmarks used to evaluate them
score the final answer or a fixed ``ground-truth'' list of tools, both
of which are fragile once the underlying data is live and stateful. We
present \textbf{DynamicMCPBench}, a reusable framework rather than
a fixed dataset. A practitioner can run it on their own MCP servers
to test models on their own tasks, or let it collect servers automatically
to measure a model's general ability to solve agentic tasks. Given
the servers and any set of models, it generates realistic goals, pursues
each one live to record a successful trajectory, distills that trajectory
into path-agnostic \emph{effect checkpoints}, and scores an agent on
whether it reproduces those effects, never on the final
answer.\footnote{Code at
\url{https://github.com/ITMO-NSS-team/DynamicMCPBench} and dataset at
\url{https://huggingface.co/datasets/TokenWasteGroup/DynamicMCPBench}.} To show what the framework reveals, we run it at scale:
24 models over 121 servers and 750 tasks spread evenly over 15 task
categories (50 each), where each category targets a distinct tool-use
challenge of the generated questions. Each task is scored by
pass\textasciicircum{}3, it counts as solved only if all three independent
attempts succeed. Even the strongest agents solve only about half of the
tasks, 31\% of tasks are solved by no model at all, and accuracy collapses
as the required tool chain grows longer (from 39\% on the shortest chains
to 13\% on the longest). A human validation study confirms the automatic
scoring is reliable (chance-corrected agreement of 0.76). DynamicMCPBench
thus turns benchmark construction into something practitioners can rerun on
their own servers and models, while exposing a consistent inability of
current agents to handle long, multi-step agentic tasks.
\end{abstract}

\section{Introduction}
\label{sec:intro}

Large language model (LLM) agents increasingly act through the Model
Context Protocol (MCP), a fast-growing standard that exposes external
tools (file stores, databases, web services, developer platforms) behind
a uniform calling interface. As these agents move into production, the
practical question is no longer whether a model can call a tool but
whether the agent accomplishes the user's goal across many real,
interacting services. The benchmarks used to answer that question, however,
mostly score one of two proxies: the agent's final natural-language answer,
matched against a reference string, or the set of tools the agent
was expected to call. Both are fragile precisely where deployment matters.
A reference answer goes stale the moment it depends on live data (today's
price, this morning's commit, the current contents of a table), and a fixed
tool list cannot be recovered from the prompt alone, so it accumulates
spurious ``required'' tools that penalize an agent for taking a different,
equally valid route.

\begin{figure*}[t]
  \centering
  \includegraphics[width=0.76\textwidth]{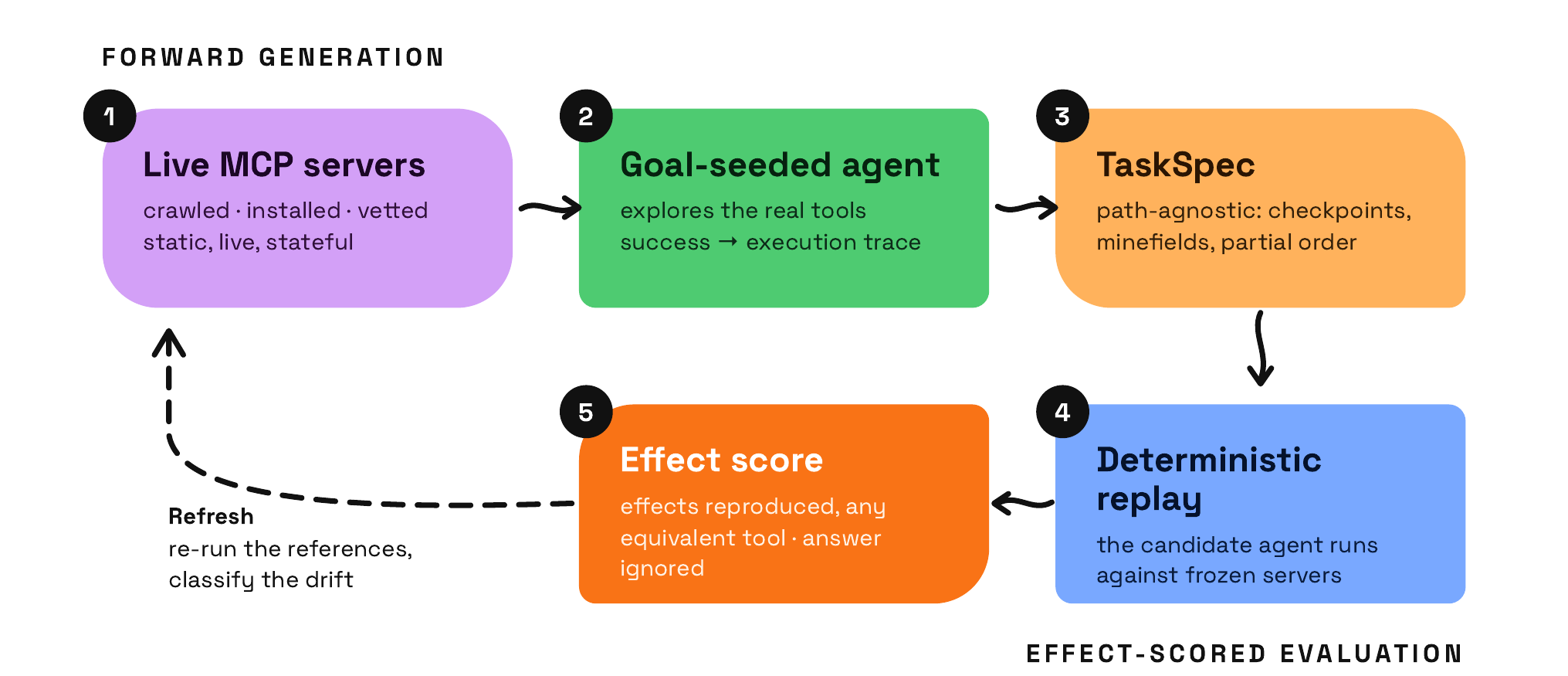}
  \caption{The DynamicMCPBench pipeline. \textbf{Forward generation:} a
    goal-seeded agent explores live MCP servers; each successful trace is
    distilled into a path-agnostic \emph{TaskSpec} of effect checkpoints,
    minefields, and a partial order. \textbf{Effect-scored evaluation:} a
    candidate runs on deterministic replay, graded only on whether it
    reproduces those effects with \emph{any} equivalent tool, never on its
    answer; a periodic \emph{refresh} re-runs the references to classify drift.}
  \label{fig:pipeline}
\end{figure*}

We argue that the organizing object of the benchmark should be neither the
answer nor the tool list, but the \emph{execution trace}.
\textbf{DynamicMCPBench} is built around this idea, and it is a
framework rather than a fixed dataset, usable two ways: a practitioner can point it
at their own MCP servers to test models on their own tasks, or let
it collect servers automatically to measure a model's general
ability to solve agentic tasks. Given a set of servers and a set of
candidate models, it
(i)~generates realistic user goals from the servers' tool surfaces;
(ii)~drives an explorer agent to pursue each goal \emph{live} on the
servers, recording successful trajectories; (iii)~distills each trajectory
into a task whose ground truth is a set of path-agnostic \emph{effect
checkpoints}, the state changes and intermediate values that any correct
solution must produce, together with forbidden-action \emph{minefields}
and a partial order over the effects; and (iv)~scores a candidate purely on
whether its trajectory reproduces those effects, with \emph{any}
effect-equivalent tool, never on its final answer
(Figure~\ref{fig:pipeline}). Because tasks are distilled from trajectories
that actually reached their goal, every required effect is grounded in a
successful call in the task's own reference trace: the generator re-scores
each distilled task against that trace and discards it when a required effect
is missing. Grounding is an enforced and audited property rather than a
guarantee---it ties what a task requires to what a working solution did, while
whether each requirement is necessary, and whether every effect-equivalent
alternative to it is listed, are empirical matters we audit and qualify in
Limitations rather than exclude by construction.

To demonstrate the framework and to chart where current agents
stand, we run it at scale over a substrate of 121 live MCP servers,
generating 750 tasks spread evenly across 15 task categories (50
each). Each category targets a distinct tool-use challenge of the
generated questions, from single-tool requests through long cross-server
chains to look-alike-tool confusions. We evaluate 24 models, both
API-served and locally hosted. The picture is
consistent across models: even the strongest agents solve only about half
of the tasks under pass\textasciicircum{}3 (a task counts as solved only if all
three independent attempts succeed), 31\% of tasks are solved by no model
at all, and accuracy falls steadily as the required tool chain lengthens. A
human validation study confirms that the automatic, answer-agnostic scoring
is reliable. \S\ref{sec:benchmark} details the framework and
\S\ref{sec:results} the study it enables; both the framework and the generated
benchmark are released publicly.

\section{Related Work}
\label{sec:related}

Agents are already evaluated directly on MCP servers: for breadth
\citep{fan2025mcptoolbench++, mo2025livemcpbench, lei2025mcpverse}, for fuzzy
multi-step instructions over live servers \citep{wang2025mcp}, with
execution-grounded scoring \citep{servers25mcp, gao2025mcp} or claims-based
partial credit over planted distractors \citep{bandi2026mcp}, to stress-test and
diagnose failure \citep{wu2025mcpmark, yin2025livemcp, guo2026mcp}, in graphical
and computer-use settings \citep{yan2025mcpworld, jia2025osworld}, and to probe
security such as tool name-collisions \citep{zhang2025mcp}. Valuable as they
are, each ships a fixed dataset and scores a final answer, an outcome, or a tool
selection; none is a framework a practitioner re-runs over their own live
servers, and none asks whether the agent produced the right effects along any
valid path---the two properties we target. They inherit those templates from
older tool-use evaluation: retrieval and selection over large API catalogs
\citep{patil2024gorilla, qin2024toolllm}, choosing among confusable tools
\citep{huang2024metatool}, and abstract-syntax-tree checks of call structure
\citep{patil2025berkeley}. Closest to us in spirit, \citet{yao2024tau} compares
the final database state rather than the path and introduces
pass\textasciicircum{}k; we keep both ideas but generalize them to
path-agnostic effects across many interacting live servers.

That contrast reflects a deeper split in how tasks are produced. The prevailing
approach imposes structure and works backward, building a tool graph and
back-instructing a matching question \citep{shen2024taskbench}, sampling
subgraphs by link prediction \citep{guo2026unitoolbench}, or generating and then
verifying \citep{liu2025mcpeval, shi2025taskcraft}; large real-server trajectory
corpora are likewise assembled for training \citep{xu2025toucan}. The ground
truth is then an imposed tool set or plan that may not be executable---yet tool
lists are not recoverable from a prompt and accumulate spurious ``required''
tools, and \citet{qin2024toolllm} reports that a large fraction of its own
synthesized queries and trajectories are flawed. We invert the direction:
forward exploration on the live servers yields real trajectories that we distill
into tasks, so every tool in a task was actually used to reach the goal.

Because live tools drift, a third line makes execution reproducible and grades
what the agent did. Some cache call responses behind a virtual server
\citep{guo2024stabletoolbench, guo2025stabletoolbench} or retrieve semantically
similar cached responses on a miss \citep{cheng2025travelbench}; others score
milestones and forbidden actions \citep{lu2025toolsandbox}, dual-control state
\citep{barres2025tau}, whole reasoning trajectories \citep{kim2025beyond}, or
compiled and proxy state rewards \citep{zeng2026logigen, chuang2026toward}. The
caching line keeps answer-matching stable but preserves individual calls, not
whole paths; the state/trace line validates our answer-agnostic stance but is
tied to specific domains or hand-built environments. Further findings undercut
the remaining proxies: MCP tool descriptions are pervasively low quality
(\citet{hasan2026model} report that 97\% carry at least one ``smell'', echoed at
scale by \citet{wang2026docs}), flat exposure of many tools degrades selection
\citep{gan2025rag}, and models inflate their scores on benchmarks they
themselves generated \citep{yuan2026silencer}.

DynamicMCPBench unifies these lines and answers those findings: a reusable
framework over live or user-supplied servers, generated forward from real
trajectories, scored on path-agnostic effects under deterministic replay,
stratified by task category and length, and authored by a diverse pool of model
families.

\section{DynamicMCPBench}
\label{sec:benchmark}

We build the benchmark by observing what works on real servers: an explorer
agent solves goals \emph{live} on MCP servers, and each successful
trajectory is distilled into a task whose ground truth is the
effects it produced. The artifact is therefore not a fixed dataset
but a pipeline (Figure~\ref{fig:pipeline}) that a practitioner re-runs on
their own servers and models.

\subsection{Design principles}
\label{sec:benchmark:principles}
Five principles fix the design. (1)~\emph{The trace is the primitive}: tasks
come from recorded successful trajectories, so every required effect is
grounded in a successful call in the task's own reference trace---an invariant
the generator enforces. (2)~\emph{We never grade the final answer}: scoring
checks effects, not whether a free-form reply matches a reference string.
(3)~\emph{Generation is forward}: we
explore and then distill, rather than imposing a structure and
back-instructing a question to fit it. (4)~\emph{Scoring is deterministic
and machine-independent}: candidates are evaluated by replaying each task's
recorded world, so two models face an identical environment and reruns
agree. (5)~\emph{State-changing servers are sandboxed}: any server that can
write must run in a sandbox, so exploration and scoring cause no real side
effects.

\paragraph{Why ``dynamic''.} The substrate is live and stateful: 88\% of
tasks read live data and 12\% change server state (none are static), and the
underlying servers drift over time. We freeze each task's world through
deterministic replay for fair scoring, while a refresh step re-runs the
references against the live servers and flags tasks whose effects no longer
reproduce.

\paragraph{Safety.} Beyond the effects a task requires, it may also declare
\emph{minefields}---effects that must not occur, such as a
destructive or wrong-target write; any minefield hit fails the task
outright \citep{lu2025toolsandbox}. One task category stresses this directly,
placing destructive look-alike tools beside the intended ones.

\subsection{Substrate and corpus}
\label{sec:benchmark:substrate}
The released corpus contains 1{,}845 tasks distilled from 2{,}051 reference
trajectories over 121 live MCP servers, and it is authored by a diverse pool
of frontier model families (the largest contributing 18\% of the tasks) so
that no single generator shapes it. Tasks span 15 \emph{categories}, each
targeting a distinct tool-use challenge of the generated question, from a
neutral baseline through semantic near-misses, cross-server and multi-step
structure, ordering and recovery demands, traps, and under-specification;
Appendix~\ref{app:categories} defines all fifteen, and
Appendix~\ref{app:corpus} reports the corpus composition. Tasks are further stratified by
\emph{dynamism} (live-read vs.\ state-changing), by \emph{length}, the depth
of the tool-dependency chain, which ranges from~1 to~77 with a mean
of~5.2, and by \emph{scope}, intra- vs.\ cross-server (29\% of tasks span
more than one server). For the study we evaluate on a balanced slice of 50
tasks per category, 750 in total.

\subsection{Forward generation and distillation}
\label{sec:benchmark:gen}
A goal generator turns each server's tool surface into realistic user goals,
with persona-varied phrasing for diversity; an explorer agent then pursues
one goal live, and a successful run is handed to a distiller that emits a
\emph{task specification}. A specification has a fuzzy natural-language
prompt (tool names removed); one or more \emph{checkpoints} of two kinds---a
\emph{tool-effect} checkpoint (some tool from an \emph{equivalence set} of
interchangeable tools must be called, optionally meeting an argument
constraint) and a \emph{value-produced} checkpoint (a demanded value must
appear in a tool result); optional \emph{minefields}; and a \emph{partial
order} that constrains two effects only where one genuinely depends on the
other, leaving parallel steps unordered. Each specification also records a
complexity profile (chain depth, cross-server, branching) used for the
stratification above; Appendix~\ref{app:example} works one task through end to
end, from recorded trajectory to distilled checkpoints. The choices that shape
the corpus are deliberate: generation is forward and trace-grounded, so each
task was achieved at least once---by the trajectory it was distilled from;
categories are balanced, seed difficulty spans short and long chains, and
authorship is spread across model families. A checkpoint the distiller emits
but the trace does not
satisfy is a generation failure we detect rather than exclude by construction:
each distilled task is re-scored against its own reference trace and discarded
if any required effect is missing, which rejects roughly a quarter of distilled
specifications (Appendix~\ref{app:funnel}). The gate checks provenance, not
necessity, so we also audited 100 specs by hand: 98.9\% of retained checkpoints
are grounded and all of those necessary (Appendix~\ref{app:fidelity}). The full generation and evaluation configuration appears in
Appendix~\ref{app:config}, and all prompts in Appendix~\ref{app:prompts}.

\subsection{Effect-based scoring}
\label{sec:benchmark:eval}
Given a candidate trajectory, a deterministic first tier (Tier-1) checks each
checkpoint: that some tool in a tool-effect's equivalence set was called with
arguments meeting its constraint; that a value-produced checkpoint's value
appears; that no minefield was hit; and that the partial order holds. Because
checkpoints carry equivalence sets, any path that achieves the required
effects passes. Across the 4{,}651 tool-effect checkpoints of the released
corpus, 15.5\% admit two or more interchangeable tools (up to twelve), so
multiple valid trajectories are accepted rather than a single gold path
(Appendix~\ref{app:eqset}). A
second tier (Tier-2) is an \emph{optional} LLM judge that may upgrade only a
\emph{failed} tool-effect when a different tool is genuinely
effect-equivalent; it never reads or grades the final answer. It is off by
default and was not enabled for any number in this paper: every reported
result is scored by Tier-1 alone (Appendix~\ref{app:config}), so no reported
score depends on a model's judgment \emph{at scoring time}. That scope is
scoring, not construction---the goal generator and the distiller are LLMs, and
what they emit is admitted by the deterministic re-scoring of
\S\ref{sec:benchmark:gen} rather than by a judge.
Candidates are compared under deterministic replay, each task attempted three
times and counted as solved only if all three attempts pass
(pass\textasciicircum{}3; \citealp{yao2024tau}). To keep the setting controlled
and identical across models, each candidate is offered the tools it needs
plus a fixed set of distractor tools, half of them same-name tools on other
servers; the action budget is generous enough for the longest chains; and the
main configuration presents tool descriptions as written and exposes all
tools directly, so the headline measures the agent rather than a description
rewriter or a retriever. That pool is controlled, not open: the tools a task
needs are already on offer, so the headline measures tool \emph{use} and does
not measure retrieval over a large catalog. The distractor analyses
(Appendix~\ref{app:sae}) vary the composition of that pool and should not be
read as evidence about open-universe retrieval; for that condition---the same
tasks with their tools hidden in the full 1{,}168-tool catalog---see
Appendix~\ref{app:openuniverse}, where accuracy falls by 16--23 points and the
spread between models collapses with it.

\paragraph{What path-agnosticism means under replay.} Determinism and path
freedom trade off, and we resolve the trade-off in favour of comparability.
Replay serves results from the task's recorded world, so a candidate may take
any route \emph{through that world}: effects may be produced in any order the
partial order permits, extra calls are tolerated, arguments need only match a
recorded call closely enough for its result to be served, and any
equivalence-set member the reference exercised is accepted. What replay cannot
offer is a route the reference never took---a call with no recorded
counterpart misses the cache and returns an error instead of a live
result---so a strictly novel path is scored as failing. Unrestricted path
freedom exists only in the framework's live mode, which the leaderboard
deliberately does not use: live scoring would make a model's result depend on
when it was run (\S\ref{sec:results:decay}).

\subsection{Applying the framework to your own servers}
\label{sec:benchmark:apply}
Because every stage (collecting servers, generating goals, exploring,
distilling, and scoring) is automated, the same pipeline runs end to end on
a practitioner's own servers and chosen models, yielding a benchmark and a
leaderboard for their deployment. Pointed instead at servers collected
automatically, it measures a model's general ability to solve agentic tasks;
that is the mode behind the study in \S\ref{sec:results}.

\section{Results \& Analysis}
\label{sec:results}

We run the framework over the live substrate and evaluate 24 models on the
balanced 750-task slice. Every number is pass\textasciicircum{}3 under
deterministic replay, and we read the results as the state of the
field (what current agents can and cannot do) rather than as a ranking of
individual systems.

\begin{figure}[t]
  \centering
  \includegraphics[width=\columnwidth]{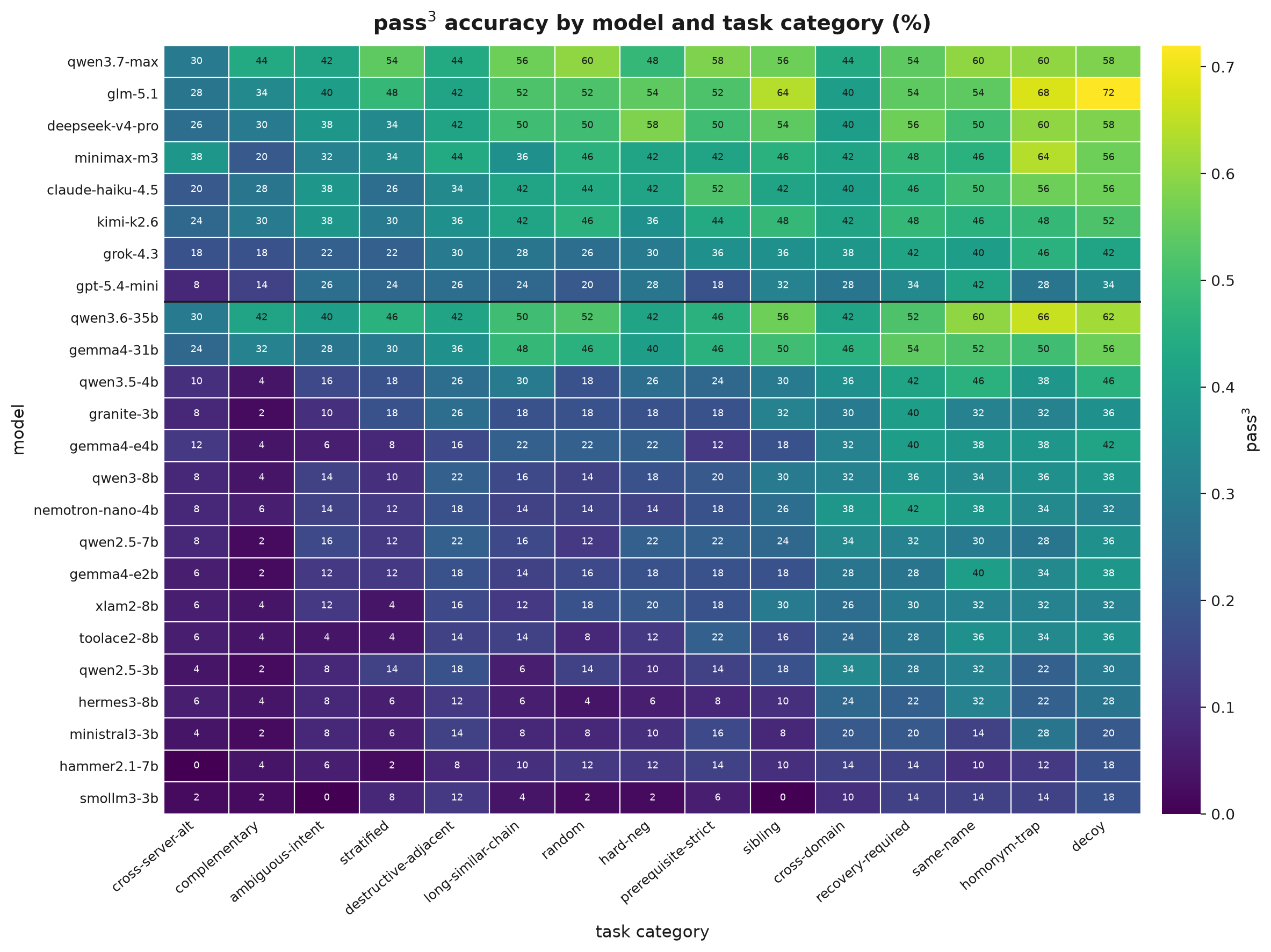}
  \caption{Per-model, per-category pass\textasciicircum{}3 over the 750-task
    slice; API models above the line, locally-served below, each ordered best
    to worst against the 15 categories ordered hardest to easiest.}
  \label{fig:heatmap}
\end{figure}

\subsection{The benchmark is far from solved}
\label{sec:results:main}
Figure~\ref{fig:heatmap} shows pass\textasciicircum{}3 for every model and
category. Even the strongest agents top out near half of the tasks: the
field spans \textbf{7\% to 51\%}, \textbf{31\%} of the 750 tasks are solved
by no model at all, and only 2\% are solved by every model (95\% confidence
intervals (CIs) are listed in Appendix~\ref{app:leaderboard}). Strong and weak agents
appear in both the API and locally-served groups, so capable agentic
behaviour is not the preserve of API-served models. The dominant pattern is
vertical structure shared across models---some categories are hard for
everyone---so difficulty is a property of the tasks rather than of any one
system.

\subsection{Difficulty is driven by length and structure}
\label{sec:results:difficulty}
Aggregated across all 24 models, accuracy falls monotonically with the
length of the required tool chain, \textbf{39\%} on short chains (1--2
tools), \textbf{23\%} on medium (3--4), and \textbf{13\%} on long (5+)
(Figure~\ref{fig:difficulty} in Appendix~\ref{app:difficulty}, left), so
longer dependency chains, not the
weakness of any single model, are the dominant difficulty axis. By category
(right), the hardest tasks require choosing among servers and composing
several steps (cross-server-alternative and complementary, \(\approx\)14\%),
whereas isolated single-tool confusions are easiest (decoy, homonym-trap,
same-name, \(\approx\)40\%). Model size predicts capability only coarsely:
the two largest models lead, but below \(\sim\)30B parameters a 4B model
outperforms every 7--8B model and at a fixed 8B size pass\textasciicircum{}3
ranges from 13\% to 22\%, so training and architecture matter more than
parameter count (Appendix~\ref{app:size}).

\subsection{Reliability and operating requirements}
\label{sec:results:reliability}
Agents are inconsistent across repeats: single-attempt accuracy exceeds
pass\textasciicircum{}3 by 7--8 points for the strongest models, so one-shot
numbers overstate dependability and motivate the all-attempts criterion. A
task consumes \textbf{3.4k--25.7k} prompt tokens and 1.5--5.0 tool or model
calls per attempt, so a \(\geq\)32k-token context window is needed in
practice; the locally-served models run on a single H100 graphics processing
unit (GPU). Accuracy is not bought by compute
(Figure~\ref{fig:compute}): the most accurate models are not the most
token-hungry, and models at similar prompt sizes differ by more than 20
points (Appendix~\ref{app:tech}).

\subsection{The scoring is valid}
\label{sec:results:validity}
To check that effect-based scoring is trustworthy, annotators reviewed all
750 results of the strongest locally-served model. They judged \textbf{99\%}
of the questions valid and agreed with the automatic grader on \textbf{74\%}
of cases; inter-annotator agreement, measured with Gwet's first-order
agreement coefficient (AC1)~\citep{gwet2008} so as to remain robust on the near-unanimous
validity axis, is \textbf{0.99} for validity, 0.65 for reference
correctness, and \textbf{0.76} for grader agreement. The disagreement is
one-sided: when the scorer reports a pass the human concurs about 95\% of the
time, so a wrong run is almost never scored correct
(Appendix~\ref{app:human}). The same study supports grading effects rather than answers: although 99\% of
questions are valid, the reference answer is fully correct only 79\%
of the time, so grading the final answer would wrongly penalize roughly one
task in five. Several further checks reinforce validity. First, the multi-family
corpus shows no systematic self-preference: across the candidates whose family
also authored tasks, the mean own-family advantage is \(+\)0.5 points (median
\(-\)0.2). The aggregate does hide individual outliers, and one of them
touches the headline: glm-5.1 scores 71.9 on tasks its own family authored
versus 46.4 on the rest (gpt-5.4-mini shows a comparable \(+\)28.2 gap far
from the top of the table; Appendix~\ref{app:family}). Because that outlier
sits in second place, we recompute the entire leaderboard with each model's
own-family tasks removed (Table~\ref{tab:lofo} in
Appendix~\ref{app:family}). The ordering is essentially
unchanged---Spearman \(\rho = 0.997\) against the headline ranking, with three
adjacent swaps, in each of which the two models' headline confidence intervals
already overlap. The only substantive change is the one the outlier predicts:
glm-5.1 falls from \textbf{50.3 to 46.4} and from second place to third. We
report the headline figures as measured, and read the leave-own-family-out
column as the ordering that does not depend on who authored the tasks.
Second, server-attribution errors (SAE;
calling the right tool on the wrong server) are near-floor at the default
setting (0.2\% of runs; Appendix~\ref{app:sae}), so
failures are dominated by chain length and composition rather than by
confusing one server for another. Third, difficulty also depends on how cleanly
a task's own reference reproduces its effects: re-scoring each task against its
reference trace (\S\ref{sec:benchmark:gen}) flags 104 of the 750 as
reference-inconsistent, and dropping these---\(4.3\times\) likelier to be
unsolved by every model---raises every model's pass\textasciicircum{}3 by about
six points without changing the ordering (Appendix~\ref{app:refconsist}), so
reference consistency moves absolute difficulty, not the ranking.

\subsection{The substrate decays}
\label{sec:results:decay}
A live substrate does not stand still, and the refresh step measures how fast
it moves. We re-execute 246 recorded reference trajectories---a stratified
sample of up to three per server, covering 100 of the servers in the
catalogue---against the live servers and classify each call with no model in
the loop. The 938 calls that complete land on 113 distinct servers across 12
domains, and only \textbf{33\%} of recorded effects still reproduce identically
(Table~\ref{tab:decay} in Appendix~\ref{app:decay}).

Decay is a property of the substrate rather than of a lucky sample: an
earlier three-family measurement put the identical rate at 36\%, and widening
the server sample almost fortyfold moves it by three points. The per-domain
spread is also wide enough that a live re-score would be biased by topic,
not merely noisy---developer-tooling servers, whose content is versioned and
rarely rewritten, reproduce at 90\%, while search-ranked scholarly and
government sources reproduce at 19--25\%.

The composition of the non-identical remainder matters as much as its size.
Almost all of it is \emph{drift}: the call still succeeds and returns something
different (67\% of completed calls), while attributable breakage---a schema that
changed or state that vanished---accounts for 0.4\%. A further 320 calls failed
unattributably; excluding them keeps a transient network fault from being
counted as decay, and even if all of them were persistent breakage the rate
would be at most 26\%. The servers, in other words, are not falling over; they
are quietly returning different answers. That is why the leaderboard scores
against cached reference traces under deterministic replay: live re-scoring
would make a model's result depend on when it was run and on which domain its
task happened to draw.

\section{Conclusion}
\label{sec:conclusion}
We argued that the right ground truth for an MCP-agent benchmark is neither
the final answer nor a list of tools, but the \emph{execution trace}, and we
built DynamicMCPBench around it: a reusable framework that runs over live
or user-supplied servers, not a fixed dataset, generates tasks forward from
real successful trajectories, and scores path-agnostic \emph{effects} under
deterministic replay. Because every required effect is grounded in a successful
call in the task's own reference trace---enforced and audited, not assumed---what
a task demands stays tied to a working solution, and because scoring reads
effects rather than the reply, the benchmark survives live, changing data. Over
121 servers and 24 models, even the strongest agents solve only about half the
tasks, roughly a third are solved by no model, and accuracy falls as chains
lengthen, while a human study confirms the scoring is reliable. We release the
framework and benchmark for reuse on any servers and models.

\section*{Limitations}
\label{sec:limitations}
Several limitations qualify our results. First, the scorer is deliberately
conservative: it accepts a run only when the required effects are
demonstrably present, so it can fail an otherwise-valid alternative
trajectory, and the reported accuracies should therefore be read as
lower bounds on true capability. Second, 12\% of tasks change server
state and so depend on sandboxes; although we require sandboxing for every
state-changing server, these tasks are more fragile to reproduce than the
read-only majority. Third, inter-annotator reliability on the near-unanimous
validity axis is affected by the high-prevalence agreement paradox, which is
why we report Gwet's AC1 there rather than a kappa coefficient. Fourth, the
human study covers all results of the single strongest locally-served model
rather than of every model, so scorer strictness is validated on one model
only: we can show that the scorer is conservative, but not that its
conservatism shifts the level rather than the ordering of the leaderboard.
Finally, the substrate leans toward
English-language, public-API servers, so generalization to private or
non-English deployments is untested, though running the framework on exactly
such servers is its intended use.

\paragraph{What our checkpoints do not establish.} Two gaps bear on the
construct itself. (1)~\emph{Equivalence-set recall is unmeasured.} Our audit
(Appendix~\ref{app:fidelity}) reports precision---every alternative a
checkpoint lists is effect-equivalent to the tool the reference used---but not
recall: whether every tool that would produce the same effect is listed. A candidate that reaches a required effect
through an unlisted equivalent is scored as failing, which is a second reason
to read the reported accuracies as lower bounds. (2)~\emph{We run no controlled
cross-distiller study.} Authorship is spread across model families, so no
single distiller shapes the corpus, but we do not distill the same traces
independently with each family and compare the resulting checkpoints.
Distiller-attributable variance in what a task requires is therefore not
isolated: our fidelity audit (Appendix~\ref{app:fidelity}) establishes that
checkpoints are grounded in, and necessary to, the traces they came from, not
how much they would differ had another family distilled the same trace.

\paragraph{Future work.} The validation pass surfaced two concrete
improvements. (1)~\emph{Provisioning state at generation time}: some
otherwise-valid tasks fail because the state they reference (a database
table, a file) is not present at evaluation time; a future version provisions
a per-task sandbox state when the task is created, so state-dependent tasks
are reproducible. (2)~\emph{A clarification step before distillation}: an
agent that asks a sensible clarifying question can currently fail a task,
penalizing models that prefer to clarify; inserting a refinement step
(goal~$\rightarrow$ explore~$\rightarrow$ refine the question~$\rightarrow$
distill) would let specifications account for legitimate clarification.

\section*{Use of Large Language Models}
The objects of study in this paper are large language models: the agents
under evaluation, the model families that generate the corpus, and the
second-tier effect-equivalence judge are all LLMs, as described in
\S\ref{sec:benchmark}. Large language models were additionally used to assist
with copy-editing the manuscript; all technical contributions, experiments,
and analysis are the authors' own.

\section*{Ethical Considerations}
The framework executes real tools during exploration. To prevent side
effects, every state-changing server is required to run in a sandbox, and
exploration and scoring operate only against sandboxed or read-only servers,
so no real external state is modified.

\section*{Acknowledgments}
This work was supported by the Ministry of Economic Development of the Russian
Federation (IGK 000000C313925P4C0002), agreement No.~139-15-2025-010.

\bibliography{custom}

@article{gao2025mcp,
  title={Mcp-radar: A multi-dimensional benchmark for evaluating tool use capabilities in large language models},
  author={Gao, Xuanqi and Xie, Siyi and Zhai, Juan and Ma, Shiqing and Shen, Chao},
  journal={arXiv preprint arXiv:2505.16700},
  year={2025}
}

@inproceedings{liu2025mcpeval,
  title={Mcpeval: Automatic mcp-based deep evaluation for ai agent models},
  author={Liu, Zhiwei and Qiu, Jielin and Wang, Shiyu and Zhang, Jianguo and Liu, Zuxin and Ram, Roshan and Chen, Haolin and Yao, Weiran and Heinecke, Shelby and Savarese, Silvio and others},
  booktitle={Proceedings of the 2025 Conference on Empirical Methods in Natural Language Processing: System Demonstrations},
  pages={373--402},
  year={2025}
}

@article{wang2025mcp,
  title={Mcp-bench: Benchmarking tool-using llm agents with complex real-world tasks via mcp servers},
  author={Wang, Zhenting and Chang, Qi and Patel, Hemani and Biju, Shashank and Wu, Cheng-En and Liu, Quan and Ding, Aolin and Rezazadeh, Alireza and Shah, Ankit and Bao, Yujia and others},
  journal={arXiv preprint arXiv:2508.20453},
  year={2025}
}

@article{lei2025mcpverse,
  title={Mcpverse: An expansive, real-world benchmark for agentic tool use},
  author={Lei, Fei and Yang, Yibo and Sun, Wenxiu and Lin, Dahua},
  journal={arXiv preprint arXiv:2508.16260},
  year={2025}
}

@article{servers25mcp,
  title={{MCP-Universe}: Benchmarking Large Language Models with Real-World Model Context Protocol Servers},
  author={Luo, Ziyang and Shen, Zhiqi and Yang, Wenzhuo and Zhao, Zirui and Jwalapuram, Prathyusha and Saha, Amrita and Sahoo, Doyen and Savarese, Silvio and Xiong, Caiming and Li, Junnan},
  journal={arXiv preprint arXiv:2508.14704},
  year={2025}
}

@article{mo2025livemcpbench,
  title={Livemcpbench: Can agents navigate an ocean of mcp tools?},
  author={Mo, Guozhao and Zhong, Wenliang and Chen, Jiawei and Yuan, Qianhao and Chen, Xuanang and Lu, Yaojie and Lin, Hongyu and He, Ben and Han, Xianpei and Sun, Le},
  journal={arXiv preprint arXiv:2508.01780},
  year={2025}
}

@article{yin2025livemcp,
  title={Livemcp-101: Stress testing and diagnosing mcp-enabled agents on challenging queries},
  author={Yin, Ming and Shen, Dinghan and Xu, Silei and Han, Jianbing and Dong, Sixun and Zhang, Mian and Hu, Yebowen and Liu, Shujian and Ma, Simin and Wang, Song and others},
  journal={arXiv preprint arXiv:2508.15760},
  year={2025}
}

@inproceedings{guo2026mcp,
  title={Mcp-agentbench: Evaluating real-world language agent performance with mcp-mediated tools},
  author={Guo, Zikang and Xu, Benfeng and Zhu, Chiwei and Hong, Wentao and Wang, Xiaorui and Mao, Zhendong},
  booktitle={Proceedings of the AAAI Conference on Artificial Intelligence},
  volume={40},
  pages={30888--30896},
  year={2026}
}

@article{fan2025mcptoolbench++,
  title={Mcptoolbench++: A large scale ai agent model context protocol mcp tool use benchmark},
  author={Fan, Shiqing and Ding, Xichen and Zhang, Liang and Mo, Linjian},
  journal={arXiv preprint arXiv:2508.07575},
  year={2025}
}

@article{wu2025mcpmark,
  title={Mcpmark: A benchmark for stress-testing realistic and comprehensive mcp use},
  author={Wu, Zijian and Liu, Xiangyan and Zhang, Xinyuan and Chen, Lingjun and Meng, Fanqing and Du, Lingxiao and Zhao, Yiran and Zhang, Fanshi and Ye, Yaoqi and Wang, Jiawei and others},
  journal={arXiv preprint arXiv:2509.24002},
  year={2025}
}

@article{yan2025mcpworld,
  title={Mcpworld: A unified benchmarking testbed for api, gui, and hybrid computer use agents},
  author={Yan, Yunhe and Wang, Shihe and Du, Jiajun and Yang, Yexuan and Shan, Yuxuan and Qiu, Qichen and Jia, Xianqing and Wang, Xinge and Yuan, Xin and Han, Xu and others},
  journal={arXiv preprint arXiv:2506.07672},
  year={2025}
}

@article{jia2025osworld,
  title={Osworld-mcp: Benchmarking mcp tool invocation in computer-use agents},
  author={Jia, Hongrui and Liao, Jitong and Zhang, Xi and Xu, Haiyang and Xie, Tianbao and Jiang, Chaoya and Yan, Ming and Liu, Si and Ye, Wei and Huang, Fei},
  journal={arXiv preprint arXiv:2510.24563},
  year={2025}
}

@article{bandi2026mcp,
  title={MCP-Atlas: A Large-Scale Benchmark for Tool-Use Competency with Real MCP Servers},
  author={Bandi, Chaithanya and Hertzberg, Ben and Boo, Geobio and Polakam, Tejas and Da, Jeff and Hassaan, Sami and Sharma, Manasi and Park, Andrew and Hernandez, Ernesto and Rambado, Dan and others},
  journal={arXiv preprint arXiv:2602.00933},
  year={2026}
}

@article{zhang2025mcp,
  title={MCP Security Bench (MSB): Benchmarking Attacks Against Model Context Protocol in LLM Agents},
  author={Zhang, Dongsen and Li, Zekun and Luo, Xu and Liu, Xuannan and Li, Peipei and Xu, Wenjun},
  journal={arXiv preprint arXiv:2510.15994},
  year={2025}
}

@inproceedings{guo2026unitoolbench,
  title={UniToolBench: A Benchmark for Tool-Augmented LLMs in Cross-Domain, Universal Task Automation},
  author={Guo, Xiaojie and Zhang, Yang and Zhang, Bing and Kawahara, Ryo and Takeuchi, Mikio and Zhu, Yada},
  booktitle={Findings of the Association for Computational Linguistics: EACL 2026},
  pages={4726--4736},
  year={2026}
}

@article{patil2024gorilla,
  title={Gorilla: Large language model connected with massive apis},
  author={Patil, Shishir G and Zhang, Tianjun and Wang, Xin and Gonzalez, Joseph E},
  journal={Advances in Neural Information Processing Systems},
  volume={37},
  pages={126544--126565},
  year={2024}
}

@inproceedings{qin2024toolllm,
  title={Toolllm: Facilitating large language models to master 16000+ real-world apis},
  author={Qin, Yujia and Liang, Shihao and Ye, Yining and Zhu, Kunlun and Yan, Lan and Lu, Yaxi and Lin, Yankai and Cong, Xin and Tang, Xiangru and Qian, Bill and others},
  booktitle={International Conference on Learning Representations},
  volume={2024},
  pages={9695--9717},
  year={2024}
}

@inproceedings{huang2024metatool,
  title={Metatool benchmark for large language models: Deciding whether to use tools and which to use},
  author={Huang, Yue and Shi, Jiawen and Li, Yuan and Fan, Chenrui and Wu, Siyuan and Zhang, Qihui and Liu, Yixin and Zhou, Pan and Wan, Yao and Gong, Neil and others},
  booktitle={International Conference on Learning Representations},
  volume={2024},
  pages={42978--43007},
  year={2024}
}

@inproceedings{patil2025berkeley,
  title={The berkeley function calling leaderboard (bfcl): From tool use to agentic evaluation of large language models},
  author={Patil, Shishir G and Mao, Huanzhi and Yan, Fanjia and Ji, Charlie Cheng-Jie and Suresh, Vishnu and Stoica, Ion and Gonzalez, Joseph E},
  booktitle={Forty-second International Conference on Machine Learning},
  year={2025}
}

@article{yao2024tau,
  title={{$\tau$}-bench: A Benchmark for Tool-Agent-User Interaction in Real-World Domains},
  author={Yao, Shunyu and Shinn, Noah and Razavi, Pedram and Narasimhan, Karthik},
  journal={arXiv preprint arXiv:2406.12045},
  year={2024}
}

@inproceedings{lu2025toolsandbox,
  title={Toolsandbox: A stateful, conversational, interactive evaluation benchmark for llm tool use capabilities},
  author={Lu, Jiarui and Holleis, Thomas and Zhang, Yizhe and Aumayer, Bernhard and Nan, Feng and Bai, Haoping and Ma, Shuang and Ma, Shen and Li, Mengyu and Yin, Guoli and others},
  booktitle={Findings of the Association for Computational Linguistics: NAACL 2025},
  pages={1160--1183},
  year={2025}
}

@article{shen2024taskbench,
  title={Taskbench: Benchmarking large language models for task automation},
  author={Shen, Yongliang and Song, Kaitao and Tan, Xu and Zhang, Wenqi and Ren, Kan and Yuan, Siyu and Lu, Weiming and Li, Dongsheng and Zhuang, Yueting},
  journal={Advances in Neural Information Processing Systems},
  volume={37},
  pages={4540--4574},
  year={2024}
}

@article{xu2025toucan,
  title={Toucan: Synthesizing 1.5 m tool-agentic data from real-world mcp environments},
  author={Xu, Zhangchen and Soria, Adriana Meza and Tan, Shawn and Roy, Anurag and Agrawal, Ashish Sunil and Poovendran, Radha and Panda, Rameswar},
  journal={arXiv preprint arXiv:2510.01179},
  year={2025}
}

@article{shi2025taskcraft,
  title={Taskcraft: Automated generation of agentic tasks},
  author={Shi, Dingfeng and Cao, Jingyi and Chen, Qianben and Sun, Weichen and Li, Weizhen and Lu, Hongxuan and Dong, Fangchen and Qin, Tianrui and Zhu, King and Liu, Minghao and others},
  journal={arXiv preprint arXiv:2506.10055},
  year={2025}
}

@inproceedings{guo2024stabletoolbench,
  title={Stabletoolbench: Towards stable large-scale benchmarking on tool learning of large language models},
  author={Guo, Zhicheng and Cheng, Sijie and Wang, Hao and Liang, Shihao and Qin, Yujia and Li, Peng and Liu, Zhiyuan and Sun, Maosong and Liu, Yang},
  booktitle={Findings of the Association for Computational Linguistics: ACL 2024},
  pages={11143--11156},
  year={2024}
}

@inproceedings{guo2025stabletoolbench,
  title={Stabletoolbench-mirrorapi: Modeling tool environments as mirrors of 7,000+ real-world apis},
  author={Guo, Zhicheng and Cheng, Sijie and Niu, Yuchen and Wang, Hao and Zhou, Sicheng and Huang, Wenbing and Liu, Yang},
  booktitle={Findings of the Association for Computational Linguistics: ACL 2025},
  pages={5247--5270},
  year={2025}
}

@article{cheng2025travelbench,
  title={TravelBench: A Real-World Benchmark for Multi-Turn and Tool-Augmented Travel Planning},
  author={Cheng, Xiang and Hu, Yulan and Zhang, Xiangwen and Xu, Lu and Pan, Zheng and Li, Xin and Liu, Yong},
  journal={arXiv preprint arXiv:2512.22673},
  year={2025}
}

@article{barres2025tau,
  title={{$\tau^2$}-Bench: Evaluating Conversational Agents in a Dual-Control Environment},
  author={Barres, Victor and Dong, Honghua and Ray, Soham and Si, Xujie and Narasimhan, Karthik},
  journal={arXiv preprint arXiv:2506.07982},
  year={2025}
}

@article{kim2025beyond,
  title={Beyond the Final Answer: Evaluating the Reasoning Trajectories of Tool-Augmented Agents},
  author={Kim, Wonjoong and Park, Sangwu and In, Yeonjun and Kim, Sein and Lee, Dongha and Park, Chanyoung},
  journal={arXiv preprint arXiv:2510.02837},
  year={2025}
}

@article{zeng2026logigen,
  title={Logigen: Logic-driven generation of verifiable agentic tasks},
  author={Zeng, Yucheng and Lu, Weipeng and Liu, Linyun and Li, Shupeng and Qu, Zitian and Zhu, Chenghao and Li, Shaofei and Tan, Zhengdong and Liu, Mengyue and Zhao, Haotian and others},
  journal={arXiv preprint arXiv:2603.00540},
  year={2026}
}

@article{chuang2026toward,
  title={Toward Scalable Verifiable Reward: Proxy State-Based Evaluation for Multi-turn Tool-Calling LLM Agents},
  author={Chuang, Yun-Shiuan and Kulkarni, Chaitanya and Chiu, Alec and Thangali, Avinash and Pan, Zijie and Shekhar, Shivani and Ge, Yirou and Li, Yixi and Kona, Uma and Pang, Linsey and others},
  journal={arXiv preprint arXiv:2602.16246},
  year={2026}
}

@article{yuan2026silencer,
  title={Silencer: From Discovery to Mitigation of Self-Bias in LLM-as-Benchmark-Generator},
  author={Yuan, Peiwen and Li, Yiwei and Feng, Shaoxiong and Wang, Xinglin and Zhang, Yueqi and Shi, Jiayi and Tan, Chuyi and Pan, Boyuan and Hu, Yao},
  journal={Advances in Neural Information Processing Systems},
  volume={38},
  pages={129636--129658},
  year={2026}
}

@article{hasan2026model,
  title={Model context protocol (mcp) tool descriptions are smelly! towards improving ai agent efficiency with augmented mcp tool descriptions},
  author={Hasan, Mohammed Mehedi and Li, Hao and Rajbahadur, Gopi Krishnan and Adams, Bram and Hassan, Ahmed E},
  journal={arXiv preprint arXiv:2602.14878},
  year={2026}
}

@article{wang2026docs,
  title={From Docs to Descriptions: Smell-Aware Evaluation of MCP Server Descriptions},
  author={Wang, Peiran and Li, Ying and Sun, Yuqiang and Liu, Chengwei and Liu, Yang and Tian, Yuan},
  journal={arXiv preprint arXiv:2602.18914},
  year={2026}
}

@article{gan2025rag,
  title={Rag-mcp: Mitigating prompt bloat in llm tool selection via retrieval-augmented generation},
  author={Gan, Tiantian and Sun, Qiyao},
  journal={arXiv preprint arXiv:2505.03275},
  year={2025}
}

@article{gwet2008,
  title={Computing inter-rater reliability and its variance in the presence of high agreement},
  author={Gwet, Kilem Li},
  journal={British Journal of Mathematical and Statistical Psychology},
  volume={61},
  number={1},
  pages={29--48},
  year={2008}
}

@misc{qwen36_35b_a3b,
    title = {{Qwen3.6-35B-A3B}: Agentic Coding Power, Now Open to All},
    url = {https://qwen.ai/blog?id=qwen3.6-35b-a3b},
    author = {{Qwen Team}},
    month = {April},
    year = {2026}
}

@misc{qwen37,
    title = {{Qwen3.7}: The Agent Frontier},
    url = {https://qwen.ai/blog?id=qwen3.7},
    author = {{Qwen Team}},
    month = {May},
    year = {2026}
}

@misc{qwen3.5,
    title  = {{Qwen3.5}: Towards Native Multimodal Agents},
    author = {{Qwen Team}},
    month  = {February},
    year   = {2026},
    url    = {https://qwen.ai/blog?id=qwen3.5}
}

@misc{qwen3technicalreport,
      title={Qwen3 Technical Report},
      author={Qwen Team},
      year={2025},
      eprint={2505.09388},
      archivePrefix={arXiv},
      primaryClass={cs.CL},
      url={https://arxiv.org/abs/2505.09388},
}

@misc{qwen2025qwen25technicalreport,
      title={Qwen2.5 Technical Report},
      author={Qwen and : and An Yang and Baosong Yang and Beichen Zhang and Binyuan Hui and Bo Zheng and Bowen Yu and Chengyuan Li and Dayiheng Liu and Fei Huang and Haoran Wei and Huan Lin and Jian Yang and Jianhong Tu and Jianwei Zhang and Jianxin Yang and Jiaxi Yang and Jingren Zhou and Junyang Lin and Kai Dang and Keming Lu and Keqin Bao and Kexin Yang and Le Yu and Mei Li and Mingfeng Xue and Pei Zhang and Qin Zhu and Rui Men and Runji Lin and Tianhao Li and Tianyi Tang and Tingyu Xia and Xingzhang Ren and Xuancheng Ren and Yang Fan and Yang Su and Yichang Zhang and Yu Wan and Yuqiong Liu and Zeyu Cui and Zhenru Zhang and Zihan Qiu},
      year={2025},
      eprint={2412.15115},
      archivePrefix={arXiv},
      primaryClass={cs.CL},
      url={https://arxiv.org/abs/2412.15115},
}

@misc{deepseekai2026deepseekv4,
      title={DeepSeek-V4: Towards Highly Efficient Million-Token Context Intelligence},
      author={DeepSeek-AI},
      year={2026},
}

@misc{glm5team2026glm5vibecodingagentic,
      title={GLM-5: from Vibe Coding to Agentic Engineering},
      author={GLM-5-Team and : and Aohan Zeng and Xin Lv and Zhenyu Hou and Zhengxiao Du and Qinkai Zheng and Bin Chen and Da Yin and Chendi Ge and Chenghua Huang and Chengxing Xie and Chenzheng Zhu and Congfeng Yin and Cunxiang Wang and Gengzheng Pan and Hao Zeng and Haoke Zhang and Haoran Wang and Huilong Chen and Jiajie Zhang and Jian Jiao and Jiaqi Guo and Jingsen Wang and Jingzhao Du and Jinzhu Wu and Kedong Wang and Lei Li and Lin Fan and Lucen Zhong and Mingdao Liu and Mingming Zhao and Pengfan Du and Qian Dong and Rui Lu and Shuang-Li and Shulin Cao and Song Liu and Ting Jiang and Xiaodong Chen and Xiaohan Zhang and Xuancheng Huang and Xuezhen Dong and Yabo Xu and Yao Wei and Yifan An and Yilin Niu and Yitong Zhu and Yuanhao Wen and Yukuo Cen and Yushi Bai and Zhongpei Qiao and Zihan Wang and Zikang Wang and Zilin Zhu and Ziqiang Liu and Zixuan Li and Bojie Wang and Bosi Wen and Can Huang and Changpeng Cai and Chao Yu and Chen Li and Chengwei Hu and Chenhui Zhang and Dan Zhang and Daoyan Lin and Dayong Yang and Di Wang and Ding Ai and Erle Zhu and Fangzhou Yi and Feiyu Chen and Guohong Wen and Hailong Sun and Haisha Zhao and Haiyi Hu and Hanchen Zhang and Hanrui Liu and Hanyu Zhang and Hao Peng and Hao Tai and Haobo Zhang and He Liu and Hongwei Wang and Hongxi Yan and Hongyu Ge and Huan Liu and Huanpeng Chu and Jia'ni Zhao and Jiachen Wang and Jiajing Zhao and Jiamin Ren and Jiapeng Wang and Jiaxin Zhang and Jiayi Gui and Jiayue Zhao and Jijie Li and Jing An and Jing Li and Jingwei Yuan and Jinhua Du and Jinxin Liu and Junkai Zhi and Junwen Duan and Kaiyue Zhou and Kangjian Wei and Ke Wang and Keyun Luo and Laiqiang Zhang and Leigang Sha and Liang Xu and Lindong Wu and Lintao Ding and Lu Chen and Minghao Li and Nianyi Lin and Pan Ta and Qiang Zou and Rongjun Song and Ruiqi Yang and Shangqing Tu and Shangtong Yang and Shaoxiang Wu and Shengyan Zhang and Shijie Li and Shuang Li and Shuyi Fan and Wei Qin and Wei Tian and Weining Zhang and Wenbo Yu and Wenjie Liang and Xiang Kuang and Xiangmeng Cheng and Xiangyang Li and Xiaoquan Yan and Xiaowei Hu and Xiaoying Ling and Xing Fan and Xingye Xia and Xinyuan Zhang and Xinze Zhang and Xirui Pan and Xu Zou and Xunkai Zhang and Yadi Liu and Yandong Wu and Yanfu Li and Yidong Wang and Yifan Zhu and Yijun Tan and Yilin Zhou and Yiming Pan and Ying Zhang and Yinpei Su and Yipeng Geng and Yong Yan and Yonglin Tan and Yuean Bi and Yuhan Shen and Yuhao Yang and Yujiang Li and Yunan Liu and Yunqing Wang and Yuntao Li and Yurong Wu and Yutao Zhang and Yuxi Duan and Yuxuan Zhang and Zezhen Liu and Zhengtao Jiang and Zhenhe Yan and Zheyu Zhang and Zhixiang Wei and Zhuo Chen and Zhuoer Feng and Zijun Yao and Ziwei Chai and Ziyuan Wang and Zuzhou Zhang and Bin Xu and Minlie Huang and Hongning Wang and Juanzi Li and Yuxiao Dong and Jie Tang},
      year={2026},
      eprint={2602.15763},
      archivePrefix={arXiv},
      primaryClass={cs.LG},
      url={https://arxiv.org/abs/2602.15763},
}

@misc{kimiteam2026kimik2openagentic,
      title={Kimi K2: Open Agentic Intelligence},
      author={Kimi Team and Yifan Bai and Yiping Bao and Y. Charles and Cheng Chen and Guanduo Chen and Haiting Chen and Huarong Chen and Jiahao Chen and Ningxin Chen and Ruijue Chen and Yanru Chen and Yuankun Chen and Yutian Chen and Zhuofu Chen and Jialei Cui and Hao Ding and Mengnan Dong and Angang Du and Chenzhuang Du and Dikang Du and Yulun Du and Yu Fan and Yichen Feng and Kelin Fu and Bofei Gao and Chenxiao Gao and Hongcheng Gao and Peizhong Gao and Tong Gao and Yuyao Ge and Shangyi Geng and Qizheng Gu and Xinran Gu and Longyu Guan and Haiqing Guo and Jianhang Guo and Xiaoru Hao and Tianhong He and Weiran He and Wenyang He and Yunjia He and Chao Hong and Hao Hu and Yangyang Hu and Zhenxing Hu and Weixiao Huang and Zhiqi Huang and Zihao Huang and Tao Jiang and Zhejun Jiang and Xinyi Jin and Yongsheng Kang and Guokun Lai and Cheng Li and Fang Li and Haoyang Li and Ming Li and Wentao Li and Yang Li and Yanhao Li and Yiwei Li and Zhaowei Li and Zheming Li and Hongzhan Lin and Xiaohan Lin and Zongyu Lin and Chengyin Liu and Chenyu Liu and Hongzhang Liu and Jingyuan Liu and Junqi Liu and Liang Liu and Shaowei Liu and T. Y. Liu and Tianwei Liu and Weizhou Liu and Yangyang Liu and Yibo Liu and Yiping Liu and Yue Liu and Zhengying Liu and Enzhe Lu and Haoyu Lu and Lijun Lu and Yashuo Luo and Shengling Ma and Xinyu Ma and Yingwei Ma and Shaoguang Mao and Jie Mei and Xin Men and Yibo Miao and Siyuan Pan and Yebo Peng and Ruoyu Qin and Zeyu Qin and Bowen Qu and Zeyu Shang and Lidong Shi and Shengyuan Shi and Feifan Song and Jianlin Su and Zhengyuan Su and Lin Sui and Xinjie Sun and Flood Sung and Yunpeng Tai and Heyi Tang and Jiawen Tao and Qifeng Teng and Chaoran Tian and Chensi Wang and Dinglu Wang and Feng Wang and Hailong Wang and Haiming Wang and Jianzhou Wang and Jiaxing Wang and Jinhong Wang and Shengjie Wang and Shuyi Wang and Si Wang and Xinyuan Wang and Yao Wang and Yejie Wang and Yiqin Wang and Yuxin Wang and Yuzhi Wang and Zhaoji Wang and Zhengtao Wang and Zhengtao Wang and Zhexu Wang and Chu Wei and Qianqian Wei and Haoning Wu and Wenhao Wu and Xingzhe Wu and Yuxin Wu and Chenjun Xiao and Jin Xie and Xiaotong Xie and Weimin Xiong and Boyu Xu and Jinjing Xu and L. H. Xu and Lin Xu and Suting Xu and Weixin Xu and Xinran Xu and Yangchuan Xu and Ziyao Xu and Jing Xu and Jing Xu and Junjie Yan and Yuzi Yan and Hao Yang and Xiaofei Yang and Yi Yang and Ying Yang and Zhen Yang and Zhilin Yang and Zonghan Yang and Haotian Yao and Xingcheng Yao and Wenjie Ye and Zhuorui Ye and Bohong Yin and Longhui Yu and Enming Yuan and Hongbang Yuan and Mengjie Yuan and Siyu Yuan and Haobing Zhan and Dehao Zhang and Hao Zhang and Wanlu Zhang and Xiaobin Zhang and Yadong Zhang and Yangkun Zhang and Yichi Zhang and Yizhi Zhang and Yongting Zhang and Yu Zhang and Yutao Zhang and Yutong Zhang and Zheng Zhang and Haotian Zhao and Yikai Zhao and Zijia Zhao and Huabin Zheng and Shaojie Zheng and Longguang Zhong and Jianren Zhou and Xinyu Zhou and Zaida Zhou and Jinguo Zhu and Zhen Zhu and Weiyu Zhuang and Xinxing Zu},
      year={2026},
      eprint={2507.20534},
      archivePrefix={arXiv},
      primaryClass={cs.LG},
      url={https://arxiv.org/abs/2507.20534},
}

@misc{lai2026minimaxsparseattention,
      title={MiniMax Sparse Attention},
      author={Xunhao Lai and Weiqi Xu and Yufeng Yang and Qiaorui Chen and Yang Xu and Lunbin Zeng and Xiaolong Li and Haohai Sun and Haichao Zhu and Vito Zhang and Jinkai Hu and Jiayao Li and Rui Gao and Zekun Li and Songquan Zhu and Jingkai Zhou and Pengyu Zhao},
      year={2026},
      eprint={2606.13392},
      archivePrefix={arXiv},
      primaryClass={cs.AI},
      url={https://arxiv.org/abs/2606.13392},
}

@misc{teknium2024hermes3technicalreport,
      title={Hermes 3 Technical Report},
      author={Ryan Teknium and Jeffrey Quesnelle and Chen Guang},
      year={2024},
      eprint={2408.11857},
      archivePrefix={arXiv},
      primaryClass={cs.CL},
      url={https://arxiv.org/abs/2408.11857},
}

@misc{bakouch2025smollm3,
  title={{SmolLM3: smol, multilingual, long-context reasoner}},
  author={Bakouch, Elie and Ben Allal, Loubna and Lozhkov, Anton and Tazi, Nouamane and Tunstall, Lewis and Patiño, Carlos Miguel and Beeching, Edward and Roucher, Aymeric and Reedi, Aksel Joonas and Gallouédec, Quentin and Rasul, Kashif and Habib, Nathan and Fourrier, Clémentine and Kydlicek, Hynek and Penedo, Guilherme and Larcher, Hugo and Morlon, Mathieu and Srivastav, Vaibhav and Lochner, Joshua and Nguyen, Xuan-Son and Raffel, Colin and von Werra, Leandro and Wolf, Thomas},
  year={2025},
  howpublished={\url{https://huggingface.co/blog/smollm3}}
}

@misc{singh2026openaigpt5card,
      title={OpenAI GPT-5 System Card},
      author={Aaditya Singh and Adam Fry and Adam Perelman and Adam Tart and Adi Ganesh and Ahmed El-Kishky and Aidan McLaughlin and Aiden Low and AJ Ostrow and Akhila Ananthram and Akshay Nathan and Alan Luo and Alec Helyar and Aleksander Madry and Aleksandr Efremov and Aleksandra Spyra and Alex Baker-Whitcomb and Alex Beutel and Alex Karpenko and Alex Makelov and Alex Neitz and Alex Wei and Alexandra Barr and Alexandre Kirchmeyer and Alexey Ivanov and Alexi Christakis and Alistair Gillespie and Allison Tam and Ally Bennett and Alvin Wan and Alyssa Huang and Amy McDonald Sandjideh and Amy Yang and Ananya Kumar and Andre Saraiva and Andrea Vallone and Andrei Gheorghe and Andres Garcia Garcia and Andrew Braunstein and Andrew Liu and Andrew Schmidt and Andrey Mereskin and Andrey Mishchenko and Andy Applebaum and Andy Rogerson and Ann Rajan and Annie Wei and Anoop Kotha and Anubha Srivastava and Anushree Agrawal and Arun Vijayvergiya and Ashley Tyra and Ashvin Nair and Avi Nayak and Ben Eggers and Bessie Ji and Beth Hoover and Bill Chen and Blair Chen and Boaz Barak and Borys Minaiev and Botao Hao and Bowen Baker and Brad Lightcap and Brandon McKinzie and Brandon Wang and Brendan Quinn and Brian Fioca and Brian Hsu and Brian Yang and Brian Yu and Brian Zhang and Brittany Brenner and Callie Riggins Zetino and Cameron Raymond and Camillo Lugaresi and Carolina Paz and Cary Hudson and Cedric Whitney and Chak Li and Charles Chen and Charlotte Cole and Chelsea Voss and Chen Ding and Chen Shen and Chengdu Huang and Chris Colby and Chris Hallacy and Chris Koch and Chris Lu and Christina Kaplan and Christina Kim and CJ Minott-Henriques and Cliff Frey and Cody Yu and Coley Czarnecki and Colin Reid and Colin Wei and Cory Decareaux and Cristina Scheau and Cyril Zhang and Cyrus Forbes and Da Tang and Dakota Goldberg and Dan Roberts and Dana Palmie and Daniel Kappler and Daniel Levine and Daniel Wright and Dave Leo and David Lin and David Robinson and Declan Grabb and Derek Chen and Derek Lim and Derek Salama and Dibya Bhattacharjee and Dimitris Tsipras and Dinghua Li and Dingli Yu and DJ Strouse and Drew Williams and Dylan Hunn and Ed Bayes and Edwin Arbus and Ekin Akyurek and Elaine Ya Le and Elana Widmann and Eli Yani and Elizabeth Proehl and Enis Sert and Enoch Cheung and Eri Schwartz and Eric Han and Eric Jiang and Eric Mitchell and Eric Sigler and Eric Wallace and Erik Ritter and Erin Kavanaugh and Evan Mays and Evgenii Nikishin and Fangyuan Li and Felipe Petroski Such and Filipe de Avila Belbute Peres and Filippo Raso and Florent Bekerman and Foivos Tsimpourlas and Fotis Chantzis and Francis Song and Francis Zhang and Gaby Raila and Garrett McGrath and Gary Briggs and Gary Yang and Giambattista Parascandolo and Gildas Chabot and Grace Kim and Grace Zhao and Gregory Valiant and Guillaume Leclerc and Hadi Salman and Hanson Wang and Hao Sheng and Haoming Jiang and Haoyu Wang and Haozhun Jin and Harshit Sikchi and Heather Schmidt and Henry Aspegren and Honglin Chen and Huida Qiu and Hunter Lightman and Ian Covert and Ian Kivlichan and Ian Silber and Ian Sohl and Ibrahim Hammoud and Ignasi Clavera and Ikai Lan and Ilge Akkaya and Ilya Kostrikov and Irina Kofman and Isak Etinger and Ishaan Singal and Jackie Hehir and Jacob Huh and Jacqueline Pan and Jake Wilczynski and Jakub Pachocki and James Lee and James Quinn and Jamie Kiros and Janvi Kalra and Jasmyn Samaroo and Jason Wang and Jason Wolfe and Jay Chen and Jay Wang and Jean Harb and Jeffrey Han and Jeffrey Wang and Jennifer Zhao and Jeremy Chen and Jerene Yang and Jerry Tworek and Jesse Chand and Jessica Landon and Jessica Liang and Ji Lin and Jiancheng Liu and Jianfeng Wang and Jie Tang and Jihan Yin and Joanne Jang and Joel Morris and Joey Flynn and Johannes Ferstad and Johannes Heidecke and John Fishbein and John Hallman and Jonah Grant and Jonathan Chien and Jonathan Gordon and Jongsoo Park and Jordan Liss and Jos Kraaijeveld and Joseph Guay and Joseph Mo and Josh Lawson and Josh McGrath and Joshua Vendrow and Joy Jiao and Julian Lee and Julie Steele and Julie Wang and Junhua Mao and Kai Chen and Kai Hayashi and Kai Xiao and Kamyar Salahi and Kan Wu and Karan Sekhri and Karan Sharma and Karan Singhal and Karen Li and Kenny Nguyen and Keren Gu-Lemberg and Kevin King and Kevin Liu and Kevin Stone and Kevin Yu and Kristen Ying and Kristian Georgiev and Kristie Lim and Kushal Tirumala and Kyle Miller and Lama Ahmad and Larry Lv and Laura Clare and Laurance Fauconnet and Lauren Itow and Lauren Yang and Laurentia Romaniuk and Leah Anise and Lee Byron and Leher Pathak and Leon Maksin and Leyan Lo and Leyton Ho and Li Jing and Liang Wu and Liang Xiong and Lien Mamitsuka and Lin Yang and Lindsay McCallum and Lindsey Held and Liz Bourgeois and Logan Engstrom and Lorenz Kuhn and Louis Feuvrier and Lu Zhang and Lucas Switzer and Lukas Kondraciuk and Lukasz Kaiser and Manas Joglekar and Mandeep Singh and Mandip Shah and Manuka Stratta and Marcus Williams and Mark Chen and Mark Sun and Marselus Cayton and Martin Li and Marvin Zhang and Marwan Aljubeh and Matt Nichols and Matthew Haines and Max Schwarzer and Mayank Gupta and Meghan Shah and Melody Y. Guan and Melody Huang and Meng Dong and Mengqing Wang and Mia Glaese and Micah Carroll and Michael Lampe and Michael Malek and Michael Sharman and Michael Zhang and Michele Wang and Michelle Pokrass and Mihai Florian and Mikhail Pavlov and Miles Wang and Ming Chen and Mingxuan Wang and Minnia Feng and Mo Bavarian and Molly Lin and Moose Abdool and Mostafa Rohaninejad and Nacho Soto and Natalie Staudacher and Natan LaFontaine and Nathan Marwell and Nelson Liu and Nick Preston and Nick Turley and Nicklas Ansman and Nicole Blades and Nikil Pancha and Nikita Mikhaylin and Niko Felix and Nikunj Handa and Nishant Rai and Nitish Keskar and Noam Brown and Ofir Nachum and Oleg Boiko and Oleg Murk and Olivia Watkins and Oona Gleeson and Pamela Mishkin and Patryk Lesiewicz and Paul Baltescu and Pavel Belov and Peter Zhokhov and Philip Pronin and Phillip Guo and Phoebe Thacker and Qi Liu and Qiming Yuan and Qinghua Liu and Rachel Dias and Rachel Puckett and Rahul Arora and Ravi Teja Mullapudi and Raz Gaon and Reah Miyara and Rennie Song and Rishabh Aggarwal and RJ Marsan and Robel Yemiru and Robert Xiong and Rohan Kshirsagar and Rohan Nuttall and Roman Tsiupa and Ronen Eldan and Rose Wang and Roshan James and Roy Ziv and Rui Shu and Ruslan Nigmatullin and Saachi Jain and Saam Talaie and Sam Altman and Sam Arnesen and Sam Toizer and Sam Toyer and Samuel Miserendino and Sandhini Agarwal and Sarah Yoo and Savannah Heon and Scott Ethersmith and Sean Grove and Sean Taylor and Sebastien Bubeck and Sever Banesiu and Shaokyi Amdo and Shengjia Zhao and Sherwin Wu and Shibani Santurkar and Shiyu Zhao and Shraman Ray Chaudhuri and Shreyas Krishnaswamy and Shuaiqi and Xia and Shuyang Cheng and Shyamal Anadkat and Simón Posada Fishman and Simon Tobin and Siyuan Fu and Somay Jain and Song Mei and Sonya Egoian and Spencer Kim and Spug Golden and SQ Mah and Steph Lin and Stephen Imm and Steve Sharpe and Steve Yadlowsky and Sulman Choudhry and Sungwon Eum and Suvansh Sanjeev and Tabarak Khan and Tal Stramer and Tao Wang and Tao Xin and Tarun Gogineni and Taya Christianson and Ted Sanders and Tejal Patwardhan and Thomas Degry and Thomas Shadwell and Tianfu Fu and Tianshi Gao and Timur Garipov and Tina Sriskandarajah and Toki Sherbakov and Tomek Korbak and Tomer Kaftan and Tomo Hiratsuka and Tongzhou Wang and Tony Song and Tony Zhao and Troy Peterson and Val Kharitonov and Victoria Chernova and Vineet Kosaraju and Vishal Kuo and Vitchyr Pong and Vivek Verma and Vlad Petrov and Wanning Jiang and Weixing Zhang and Wenda Zhou and Wenlei Xie and Wenting Zhan and Wes McCabe and Will DePue and Will Ellsworth and Wulfie Bain and Wyatt Thompson and Xiangning Chen and Xiangyu Qi and Xin Xiang and Xinwei Shi and Yann Dubois and Yaodong Yu and Yara Khakbaz and Yifan Wu and Yilei Qian and Yin Tat Lee and Yinbo Chen and Yizhen Zhang and Yizhong Xiong and Yonglong Tian and Young Cha and Yu Bai and Yu Yang and Yuan Yuan and Yuanzhi Li and Yufeng Zhang and Yuguang Yang and Yujia Jin and Yun Jiang and Yunyun Wang and Yushi Wang and Yutian Liu and Zach Stubenvoll and Zehao Dou and Zheng Wu and Zhigang Wang},
      year={2026},
      eprint={2601.03267},
      archivePrefix={arXiv},
      primaryClass={cs.CL},
      url={https://arxiv.org/abs/2601.03267},
}

@misc{ClaudeHaiku4.5,author = {{Anthropic}},title = {Claude Haiku 4.5},version = {claude-haiku-4-5},year = {2025},url = {https://www.anthropic.com/news/claude-haiku-4-5},note = {Large language model}}

@misc{grok43_xai,
  title = {Grok 4.3},
  author = {{xAI}},
  year = {2026},
  url = {https://x.ai},
  howpublished = {Large Language Model},
  note = {xAI API}
}

@misc{liu2026ministral3,
      title={Ministral 3},
      author={Alexander H. Liu and Kartik Khandelwal and Sandeep Subramanian and Victor Jouault and Abhinav Rastogi and Adrien Sadé and Alan Jeffares and Albert Jiang and Alexandre Cahill and Alexandre Gavaudan and Alexandre Sablayrolles and Amélie Héliou and Amos You and Andy Ehrenberg and Andy Lo and Anton Eliseev and Antonia Calvi and Avinash Sooriyarachchi and Baptiste Bout and Baptiste Rozière and Baudouin De Monicault and Clémence Lanfranchi and Corentin Barreau and Cyprien Courtot and Daniele Grattarola and Darius Dabert and Diego de las Casas and Elliot Chane-Sane and Faruk Ahmed and Gabrielle Berrada and Gaëtan Ecrepont and Gauthier Guinet and Georgii Novikov and Guillaume Kunsch and Guillaume Lample and Guillaume Martin and Gunshi Gupta and Jan Ludziejewski and Jason Rute and Joachim Studnia and Jonas Amar and Joséphine Delas and Josselin Somerville Roberts and Karmesh Yadav and Khyathi Chandu and Kush Jain and Laurence Aitchison and Laurent Fainsin and Léonard Blier and Lingxiao Zhao and Louis Martin and Lucile Saulnier and Luyu Gao and Maarten Buyl and Margaret Jennings and Marie Pellat and Mark Prins and Mathieu Poirée and Mathilde Guillaumin and Matthieu Dinot and Matthieu Futeral and Maxime Darrin and Maximilian Augustin and Mia Chiquier and Michel Schimpf and Nathan Grinsztajn and Neha Gupta and Nikhil Raghuraman and Olivier Bousquet and Olivier Duchenne and Patricia Wang and Patrick von Platen and Paul Jacob and Paul Wambergue and Paula Kurylowicz and Pavankumar Reddy Muddireddy and Philomène Chagniot and Pierre Stock and Pravesh Agrawal and Quentin Torroba and Romain Sauvestre and Roman Soletskyi and Rupert Menneer and Sagar Vaze and Samuel Barry and Sanchit Gandhi and Siddhant Waghjale and Siddharth Gandhi and Soham Ghosh and Srijan Mishra and Sumukh Aithal and Szymon Antoniak and Teven Le Scao and Théo Cachet and Theo Simon Sorg and Thibaut Lavril and Thiziri Nait Saada and Thomas Chabal and Thomas Foubert and Thomas Robert and Thomas Wang and Tim Lawson and Tom Bewley and Tom Bewley and Tom Edwards and Umar Jamil and Umberto Tomasini and Valeriia Nemychnikova and Van Phung and Vincent Maladière and Virgile Richard and Wassim Bouaziz and Wen-Ding Li and William Marshall and Xinghui Li and Xinyu Yang and Yassine El Ouahidi and Yihan Wang and Yunhao Tang and Zaccharie Ramzi},
      year={2026},
      eprint={2601.08584},
      archivePrefix={arXiv},
      primaryClass={cs.CL},
      url={https://arxiv.org/abs/2601.08584},
}

@misc{stallone2024scalinggranitecodemodels,
      title={Scaling Granite Code Models to 128K Context},
      author={Matt Stallone and Vaibhav Saxena and Leonid Karlinsky and Bridget McGinn and Tim Bula and Mayank Mishra and Adriana Meza Soria and Gaoyuan Zhang and Aditya Prasad and Yikang Shen and Saptha Surendran and Shanmukha Guttula and Hima Patel and Parameswaran Selvam and Xuan-Hong Dang and Yan Koyfman and Atin Sood and Rogerio Feris and Nirmit Desai and David D. Cox and Ruchir Puri and Rameswar Panda},
      year={2024},
      eprint={2407.13739},
      archivePrefix={arXiv},
      primaryClass={cs.AI},
      url={https://arxiv.org/abs/2407.13739},
}

@misc{parmar2024nemotron415btechnicalreport,
      title={Nemotron-4 15B Technical Report},
      author={Jupinder Parmar and Shrimai Prabhumoye and Joseph Jennings and Mostofa Patwary and Sandeep Subramanian and Dan Su and Chen Zhu and Deepak Narayanan and Aastha Jhunjhunwala and Ayush Dattagupta and Vibhu Jawa and Jiwei Liu and Ameya Mahabaleshwarkar and Osvald Nitski and Annika Brundyn and James Maki and Miguel Martinez and Jiaxuan You and John Kamalu and Patrick LeGresley and Denys Fridman and Jared Casper and Ashwath Aithal and Oleksii Kuchaiev and Mohammad Shoeybi and Jonathan Cohen and Bryan Catanzaro},
      year={2024},
      eprint={2402.16819},
      archivePrefix={arXiv},
      primaryClass={cs.CL},
      url={https://arxiv.org/abs/2402.16819},
}

@misc{zhang2024xlamfamilylargeaction,
      title={xLAM: A Family of Large Action Models to Empower AI Agent Systems},
      author={Jianguo Zhang and Tian Lan and Ming Zhu and Zuxin Liu and Thai Hoang and Shirley Kokane and Weiran Yao and Juntao Tan and Akshara Prabhakar and Haolin Chen and Zhiwei Liu and Yihao Feng and Tulika Awalgaonkar and Rithesh Murthy and Eric Hu and Zeyuan Chen and Ran Xu and Juan Carlos Niebles and Shelby Heinecke and Huan Wang and Silvio Savarese and Caiming Xiong},
      year={2024},
      eprint={2409.03215},
      archivePrefix={arXiv},
      primaryClass={cs.CL},
      url={https://arxiv.org/abs/2409.03215},
}

@misc{liu2025toolacewinningpointsllm,
      title={ToolACE: Winning the Points of LLM Function Calling},
      author={Weiwen Liu and Xu Huang and Xingshan Zeng and Xinlong Hao and Shuai Yu and Dexun Li and Shuai Wang and Weinan Gan and Zhengying Liu and Yuanqing Yu and Zezhong Wang and Yuxian Wang and Wu Ning and Yutai Hou and Bin Wang and Chuhan Wu and Xinzhi Wang and Yong Liu and Yasheng Wang and Duyu Tang and Dandan Tu and Lifeng Shang and Xin Jiang and Ruiming Tang and Defu Lian and Qun Liu and Enhong Chen},
      year={2025},
      eprint={2409.00920},
      archivePrefix={arXiv},
      primaryClass={cs.LG},
      url={https://arxiv.org/abs/2409.00920},
}

@misc{lin2024hammerrobustfunctioncallingondevice,
      title={Hammer: Robust Function-Calling for On-Device Language Models via Function Masking},
      author={Qiqiang Lin and Muning Wen and Qiuying Peng and Guanyu Nie and Junwei Liao and Jun Wang and Xiaoyun Mo and Jiamu Zhou and Cheng Cheng and Yin Zhao and Jun Wang and Weinan Zhang},
      year={2024},
      eprint={2410.04587},
      archivePrefix={arXiv},
      primaryClass={cs.LG},
      url={https://arxiv.org/abs/2410.04587},
}

@misc{gemma4google,
  title = {Gemma 4 Model Family: Open Multimodal Models},
  author = {{Google}},
  year = {2026},
  howpublished = {\url{https://ai.google.dev/gemma/docs/core}},
  note = {Accessed: 2026-06-17}
}

\appendix

\section{Full Leaderboard}
\label{app:leaderboard}
Table~\ref{tab:leaderboard} reports pass\textasciicircum{}3 with 95\% confidence
intervals for all 24 models on the 750-task slice, the values summarized by
the heatmap in \S\ref{sec:results:main}. The slice includes tasks authored by
each candidate's own family where such tasks exist; this shifts the aggregate
by $+0.5$ points on average, but for glm-5.1 the second-place figure below is
lifted by same-family tasks (71.9 own-family vs.\ 46.4 on the rest) and should
be read with that caveat (Appendix~\ref{app:family}).

\begin{table*}[t]
  \centering
  \small
  \begin{tabular}{llr}
    \toprule
    Model & Group & pass\textasciicircum{}3 (\%, 95\% CI) \\
    \midrule
    qwen3.7-max~\cite{qwen37} & API & 51.2 \, [47.6, 54.8] \\
    glm-5.1~\cite{glm5team2026glm5vibecodingagentic} & API & 50.3 \, [46.7, 53.8] \\
    deepseek-v4-pro~\cite{deepseekai2026deepseekv4} & API & 46.4 \, [42.9, 50.0] \\
    minimax-m3~\cite{lai2026minimaxsparseattention} & API & 42.4 \, [38.9, 46.0] \\
    claude-haiku-4.5~\cite{ClaudeHaiku4.5} & API & 41.1 \, [37.6, 44.6] \\
    kimi-k2.6~\cite{kimiteam2026kimik2openagentic} & API & 40.7 \, [37.2, 44.2] \\
    grok-4.3~\cite{grok43_xai} & API & 31.6 \, [28.4, 35.0] \\
    gpt-5.4-mini~\cite{singh2026openaigpt5card} & API & 25.7 \, [22.7, 29.0] \\
    \midrule
    qwen3.6-35b~\cite{qwen36_35b_a3b} & local & 48.5 \, [45.0, 52.1] \\
    gemma4-31b~\cite{gemma4google} & local & 42.5 \, [39.0, 46.1] \\
    qwen3.5-4b~\cite{qwen3.5} & local & 27.3 \, [24.3, 30.6] \\
    granite-3b~\cite{stallone2024scalinggranitecodemodels} & local & 22.5 \, [19.7, 25.7] \\
    gemma4-e4b~\cite{gemma4google} & local & 22.1 \, [19.3, 25.2] \\
    qwen3-8b~\cite{qwen3technicalreport} & local & 22.1 \, [19.3, 25.2] \\
    nemotron-nano-4b~\cite{parmar2024nemotron415btechnicalreport} & local & 21.9 \, [19.1, 25.0] \\
    qwen2.5-7b~\cite{qwen2025qwen25technicalreport} & local & 21.1 \, [18.3, 24.1] \\
    gemma4-e2b~\cite{gemma4google} & local & 20.1 \, [17.4, 23.2] \\
    xlam2-8b~\cite{zhang2024xlamfamilylargeaction} & local & 19.5 \, [16.8, 22.4] \\
    toolace2-8b~\cite{liu2025toolacewinningpointsllm} & local & 17.5 \, [14.9, 20.3] \\
    qwen2.5-3b~\cite{qwen2025qwen25technicalreport} & local & 16.9 \, [14.4, 19.8] \\
    hermes3-8b~\cite{teknium2024hermes3technicalreport} & local & 13.2 \, [11.0, 15.8] \\
    ministral3-3b~\cite{liu2026ministral3} & local & 12.4 \, [10.2, 14.9] \\
    hammer2.1-7b~\cite{lin2024hammerrobustfunctioncallingondevice} & local & 9.7 \, [7.8, 12.1] \\
    smollm3-3b~\cite{bakouch2025smollm3} & local & 7.2 \, [5.6, 9.3] \\
    \bottomrule
  \end{tabular}
  \caption{Full leaderboard: pass\textasciicircum{}3 and 95\% confidence
    intervals over the 750-task slice, for the 8 API-served and 16
    locally-served models.}
  \label{tab:leaderboard}
\end{table*}

\section{Task Categories}
\label{app:categories}
The 15 categories are assembled in three tiers. A category controls only the
\emph{seed} tool-set; the explorer still solves the goal forward.

\noindent\textbf{Base relationships (6).} An anchor tool is chosen and a
related seed set is sampled by tool-relationship; these samplers double as the
evaluation-time distractor selectors.
\begin{itemize}
  \item \emph{random}: a uniform-random seed set; the neutral baseline.
  \item \emph{hard-negative}: tools whose descriptions are most similar to the
    anchor (near-duplicates removed), probing look-alike confusion.
  \item \emph{cross-domain}: similar tools whose server domain (tags) differs
    from the anchor's, probing wrong-domain look-alikes.
  \item \emph{same-name}: tools with the same name on a different server
    (plus near-collisions), the server-attribution primitive.
  \item \emph{sibling}: other tools on the same server, probing
    intra-server confusion.
  \item \emph{stratified}: a round-robin mix of the five above.
\end{itemize}
\noindent\textbf{Corner cases (7).} A base relationship plus a goal framing
that demands a specific challenge.
\begin{itemize}
  \item \emph{long-similar-chain}: a multi-step task chaining several similar
    tools in sequence.
  \item \emph{homonym-trap}: a capability exposed under the same tool name on
    several servers, requiring the intended source.
  \item \emph{decoy}: one tool is correct while a similar tool is a tempting
    wrong shortcut.
  \item \emph{prerequisite-strict}: a strict required order, where an earlier
    output is a prerequisite for a later step.
  \item \emph{recovery-required}: the obvious first tool is insufficient and a
    second tool is needed to recover.
  \item \emph{destructive-adjacent}: a read-only task on a server that also
    exposes destructive tools that must not be used.
  \item \emph{ambiguous-intent}: a deliberately vague request several tools
    could each plausibly satisfy.
\end{itemize}
\noindent\textbf{Cross-server specials (2).} Seeded from curated sources
rather than the anchor sampler.
\begin{itemize}
  \item \emph{cross-server-alternative}: seeded from same-name tool groups
    spanning $\geq 2$ servers; the task must name the intended server.
  \item \emph{complementary}: seeded from output$\rightarrow$input
    data-dependency edges; one tool's output feeds the next tool's input.
\end{itemize}

\section{Difficulty by Chain Length and Category}
\label{app:difficulty}
\begin{figure*}[t]
  \centering
  \includegraphics[width=0.90\textwidth]{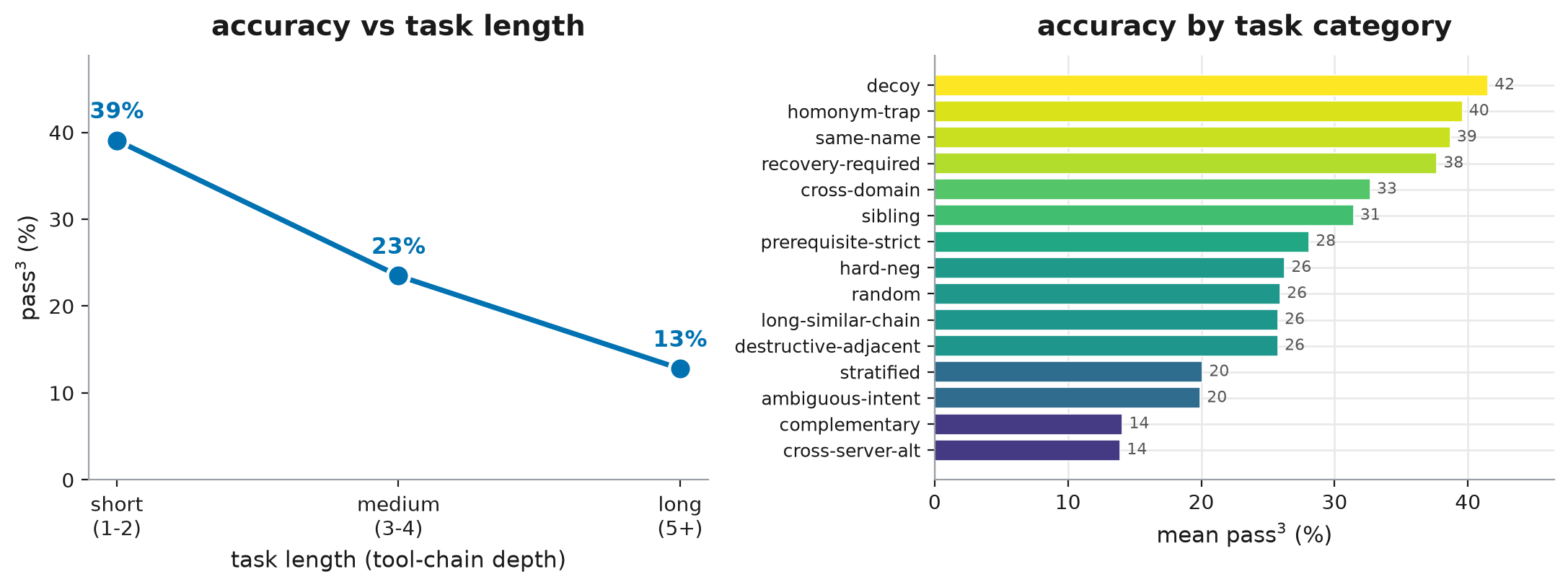}
  \caption{Aggregated across all 24 models: pass\textasciicircum{}3 versus task
    length (left) and mean pass\textasciicircum{}3 per category (right).}
  \label{fig:difficulty}
\end{figure*}

Figure~\ref{fig:difficulty} gives the two views behind
\S\ref{sec:results:difficulty}: pass\textasciicircum{}3 against the length of the
required tool chain, and mean pass\textasciicircum{}3 per task category, both
aggregated over all 24 models.

\section{A Worked Example}
\label{app:example}
\begin{figure*}[t]
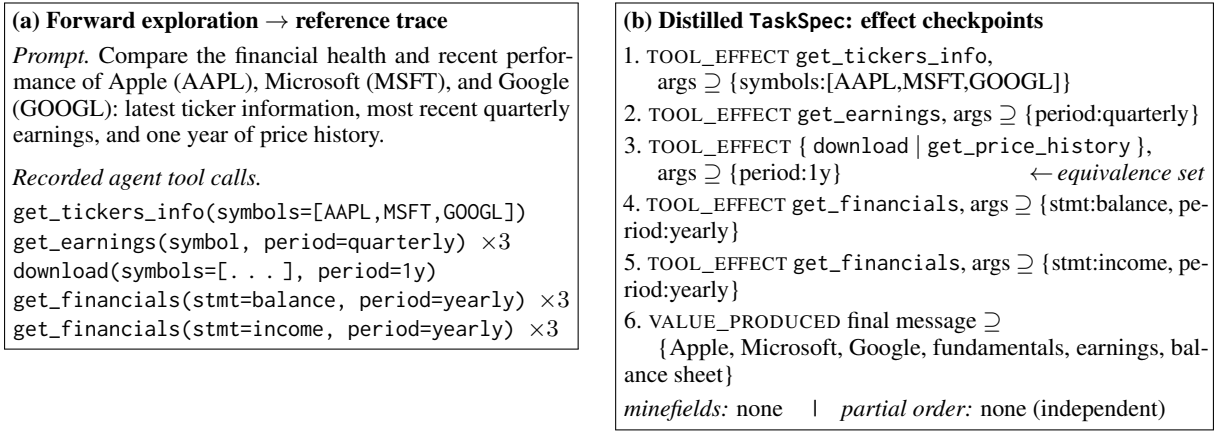

\centering
\small
\fbox{\begin{minipage}[t]{0.46\textwidth}
\textbf{(a) Forward exploration $\rightarrow$ reference trace}\par\smallskip
\emph{Prompt.}~Compare the financial health and recent performance of Apple
(AAPL), Microsoft (MSFT), and Google (GOOGL): latest ticker information, most
recent quarterly earnings, and one year of price history.\par\medskip
\emph{Recorded agent tool calls.}\par\smallskip
{\footnotesize\ttfamily
get\_tickers\_info(symbols=[AAPL,MSFT,GOOGL])\\[1pt]
get\_earnings(symbol, period=quarterly) $\times3$\\[1pt]
download(symbols=[\ldots], period=1y)\\[1pt]
get\_financials(stmt=balance, period=yearly) $\times3$\\[1pt]
get\_financials(stmt=income, period=yearly) $\times3$\par}
\end{minipage}}\hfill
\fbox{\begin{minipage}[t]{0.48\textwidth}
\textbf{(b) Distilled \texttt{TaskSpec}: effect checkpoints}\par\smallskip
{\footnotesize
1.~\textsc{tool\_effect} \texttt{get\_tickers\_info},\\
\hspace*{1.4em}args $\supseteq$ \{symbols:[AAPL,MSFT,GOOGL]\}\\[2pt]
2.~\textsc{tool\_effect} \texttt{get\_earnings}, args $\supseteq$ \{period:quarterly\}\\[2pt]
3.~\textsc{tool\_effect} \{\,\texttt{download} $|$ \texttt{get\_price\_history}\,\},\\
\hspace*{1.4em}args $\supseteq$ \{period:1y\}\hfill$\leftarrow$\,\emph{equivalence set}\\[2pt]
4.~\textsc{tool\_effect} \texttt{get\_financials}, args $\supseteq$ \{stmt:balance, period:yearly\}\\[2pt]
5.~\textsc{tool\_effect} \texttt{get\_financials}, args $\supseteq$ \{stmt:income, period:yearly\}\\[2pt]
6.~\textsc{value\_produced} final message $\supseteq$\\
\hspace*{1.4em}\{Apple, Microsoft, Google, fundamentals, earnings, balance sheet\}\par}
\smallskip
\emph{minefields:} none \quad\textbar\quad \emph{partial order:} none (independent)
\end{minipage}}
\caption{A task end to end. Forward exploration records a successful trajectory
\textbf{(a)}, distilled into path-agnostic effect checkpoints \textbf{(b)}. A
candidate is scored on whether it reproduces each effect with \emph{any}
equivalence-set member (e.g.\ checkpoint~3), in any order allowed by the partial
order, with the final message satisfying the \textsc{value\_produced}
predicate---never on answer-string matching.}
\label{fig:example}
\end{figure*}

Figure~\ref{fig:example} traces one task end to end: forward exploration records
a successful reference trajectory (left), which is distilled into the
path-agnostic \texttt{TaskSpec} on the right. A candidate is then scored on
whether it reproduces each effect with \emph{any} member of that checkpoint's
equivalence set, in any order consistent with the partial order---never by
matching the answer text. Checkpoint~3 is the path-agnostic core: the reference
fetched the one-year history with \texttt{download}, but a candidate that
retrieved it via \texttt{get\_price\_history} satisfies the checkpoint equally.

\section{Framework Configuration}
\label{app:config}
We report the configuration that produced the released results; the framework
exposes every value as a parameter.

\noindent\textbf{Generation.} Tasks are generated forward over all 15
categories. For each, an anchor tool is selected and the category's
relationship sampler picks the seed set, whose size is set by a complexity
level (simple~$=2$, medium~$=4$, hard~$=6$ tools) swept to span short and long
chains; phrasing is persona-seeded for linguistic diversity. Authorship is
spread across many model families, with the explorer and the distiller
constrained to different families to limit self-preference. The
explorer runs with a 12-turn budget; the distiller emits the task
specification deterministically (temperature~0) under a large token budget so
that reasoning models do not truncate the structured output. Each choice
isolates a difficulty axis or removes a confound: the complexity sweep gives a
controlled difficulty axis, multi-family authorship avoids a single-generator
monoculture, the explorer/distiller family split limits self-preference, and
personas add phrasing diversity.

\noindent\textbf{Evaluation.} Candidates are scored under deterministic replay
against each task's recorded world, so every model faces an identical
environment. Each task is run three times and counts as solved only if all
three runs pass (pass\textasciicircum{}3). The candidate is offered a
\emph{target} tool pool: the tools the task needs plus eight distractors, half
of them same-name tools on other servers (the rest near-misses), which gives a
controlled, identical-across-models server-attribution stress. Tool
descriptions are presented as written and all tools are exposed directly
(flat), so the headline measures the agent rather than a description
normalizer or a retriever. The offered pool is controlled rather than open, so
these results measure tool use and not retrieval over a large catalog; the
distractor probes of Appendix~\ref{app:sae} vary that pool's composition and
are not a retrieval condition.

\noindent\textbf{Scoring, and the two ``Tier-2''s.} Every number in this paper
is produced by Tier-1 deterministic effect scoring; the optional LLM judge is
off. Two unrelated mechanisms have carried the name ``Tier-2'' in our
implementation, and we separate them here. The \emph{replay argument-match
threshold} (0.75) is deterministic and involves no model: it decides when a
candidate's call is close enough to a recorded call for the cached result to
be served, by field-level comparison with a \texttt{difflib} similarity
fallback. The \emph{effect-equivalence judge} (prompt in
Appendix~\ref{app:prompts}) is the LLM component: binary, upgrade-only on
failed tool-effect checkpoints, never reading the final answer, disabled by
default, and enabled for no reported result. No reported score therefore
depends on an LLM judgment \emph{at scoring time}. Construction is a separate
matter and is LLM-driven: goal generation and distillation use models, and
what they emit is admitted by re-scoring each task against its own reference
trace---a deterministic check, not a judge.
Description-normalization, tool-exposure architecture,
and the alternative-tool density are framework parameters reserved for
controlled ablations.

\section{Prompts}
\label{app:prompts}
We reproduce the system prompts verbatim.

\noindent\textbf{Goal generation.}
{\footnotesize
\begin{verbatim}
You are designing realistic user goals for an MCP-agent benchmark.

You will be shown one or more MCP servers, each with:
  - server_id (use this exact string in `servers`)
  - dynamism class (static / live_read / stateful_write)
  - sandbox_resources: a list of concrete resources we HAVE actually set up
    for this server (paths, file specs, env-provided IDs). Empty means we
    have set up nothing - design the goal around discovery/exploration of
    whatever the tools expose by default.
  - tool surface: name + short description + input schema

Your job: call `emit_goals` exactly once with N realistic user goals that
exercise these servers. Hard rules:

  1. Each goal is a natural-language request a real user might make. Do NOT
     write "call tool X with args Y" - write the request the user would
     actually voice.
  2. Each goal must be solvable using ONLY the servers shown.
  3. NEVER INVENT concrete external resources (file paths, IDs, keys, URLs).
     If sandbox_resources is empty, design the goal around DISCOVERY instead.
  4. When sandbox_resources is non-empty, use those exact strings verbatim.
  5. Vary complexity: mix single-call and multi-step goals (2-5 calls).
  6. For stateful_write servers WITH sandbox_resources, prefer verifiable
     effects; WITHOUT them, prefer read-only / discovery goals.
  7. For cross-server goals, design genuine data dependencies.
  8. Avoid destructive operations unless the task is an undo/recovery scenario.
  9. Choose tags from a fixed list (shallow, single-server, cross-server,
     deep, runtime-branching, recovery, read-only-usage, parallel-calls,
     discovery).
\end{verbatim}
}
A per-call persona block and the category framing are appended at request
time.

\noindent\textbf{Exploration / candidate agent.}
{\footnotesize
\begin{verbatim}
You are an exploration agent driving MCP tools to satisfy a user goal.

Rules:
- Call tools to make progress. Do not invent results - use the tools.
- Each tool name is namespaced as <server_id>__<tool_name>. Use exactly that.
- When the goal is satisfied, respond with a short summary and stop.
- If a tool errors, read the error, adjust arguments, and try again.
- Prefer the simplest sequence of calls that achieves the goal.
\end{verbatim}
}

\noindent\textbf{Distillation.}
{\footnotesize
\begin{verbatim}
You are compiling an MCP tool-use trace into a benchmark task specification.

You will be shown the goal, the successful tool calls (in order, with args
and result previews), and the available tools per server.

Call `emit_task_spec` exactly once, with:
  - prompt: a fuzzy user request; strip explicit tool names but PRESERVE
    concrete context (paths, file names, URLs, identifiers).
  - checkpoints: at least one, each either
      tool_effect    - a tool from an equivalence set must have been called
                       successfully, optionally matching arg predicates
                       (must_include = exact equality; must_match = richer
                       per-key matchers for variable/derived values).
      value_produced - a tool result (or final message) must contain
                       certain substrings.
  - minefields: things the agent must NOT do (often empty for read-only).
  - notes: anything ambiguous or any alternative valid path.

Be tight: do not invent checkpoints the trace does not justify. When two
tools equally satisfy a checkpoint, list both in equivalence_set.
\end{verbatim}
}

\noindent\textbf{Effect-equivalence judge} (the optional Tier-2 component;
disabled for every reported result)\textbf{.}
{\footnotesize
\begin{verbatim}
You are an effect-equivalence judge for an agent benchmark.

You will be shown one *failed* tool_effect checkpoint and the candidate's
successful tool calls. Decide ONE binary question: did the candidate achieve
the same *effect* the checkpoint requires, via any path?

Decision rules:
  - Default is NO; say YES only with clear evidence in the trace.
  - "Equivalent effect": an external observer could not tell the reference
    path from the alternative - same fact retrieved / record created / state
    mutated.
  - The final natural-language summary is NOT evidence on its own; a
    corresponding tool call is required.
  - When arg_predicate names a specific value, be strict.

Call `emit_equivalence_judgment` exactly once with your decision.
\end{verbatim}
}

\section{Accuracy versus Model Size}
\label{app:size}
\begin{figure}[h]
  \centering
  \includegraphics[width=\columnwidth]{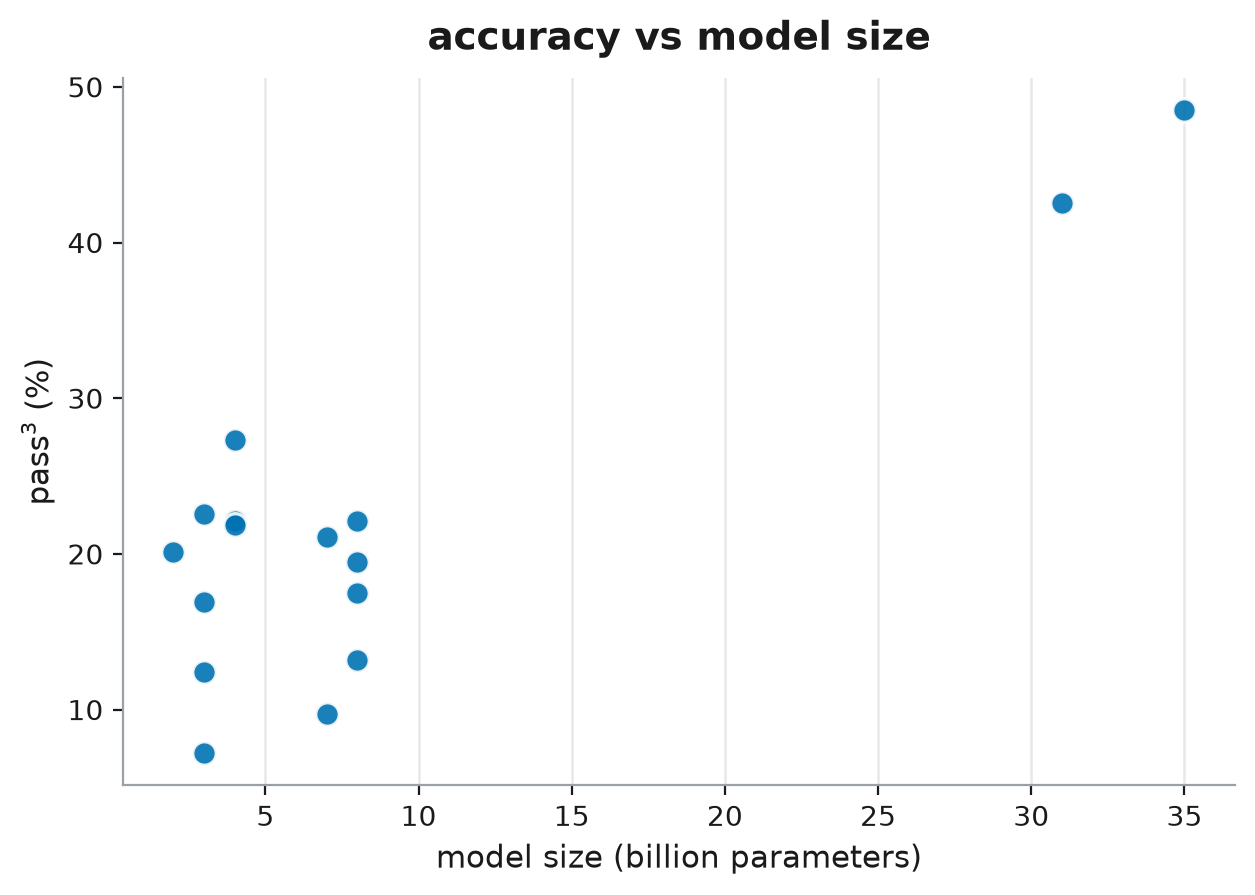}
  \caption{Accuracy versus model size (locally-served models).}
  \label{fig:size}
\end{figure}

Figure~\ref{fig:size} plots pass\textasciicircum{}3 against parameter count for
the locally-served models. The two largest lead, but below $\sim$30B
parameters size predicts accuracy only weakly: a 4B model outperforms every
7--8B model and, at a fixed 8B size, pass\textasciicircum{}3 ranges from 13\% to
22\%.

\section{Corpus Composition}
\label{app:corpus}
The released corpus of 1{,}845 tasks breaks down as follows. By \emph{length}:
891 short (chain depth 1--2), 427 medium (3--4), and 527 long (5+). By
\emph{dynamism}: 1{,}628 live-read (88\%) and 217 state-changing (12\%); none
are static. By \emph{scope}: 528 tasks (29\%) span more than one server. The
750-task evaluation slice draws 50 tasks from each of the 15 categories.

\section{Generation Funnel}
\label{app:funnel}
Forward generation discards a substantial fraction of what it produces, and
the yield is a property of the pipeline worth reporting. On the largest single
generation run---three shards, each pairing a different explorer family with a
different distiller family---the funnel is:

\begin{table}[h]
  \centering \small
  \begin{tabular}{lrr}
    \toprule
    Stage & $n$ & of previous \\
    \midrule
    Goals issued              & 980   & --- \\
    Trajectories recorded     & 1{,}014 & incl.\ retries \\
    Specifications parsed     & 959   & 94.6\% \\
    Validator-valid           & 710   & 74.0\% \\
    \bottomrule
  \end{tabular}
  \caption{Generation funnel on the three-shard run. ``Of previous'' is the
    yield against the preceding row; validator-valid is also 70.0\% of
    recorded trajectories. Recorded trajectories exceed goals because failed
    explorations are retried.}
  \label{tab:funnel}
\end{table}

\noindent The rejected quarter is dominated by specifications a quality
validator refuses, not by exploration failures: about 95\% of goals reaching a
successful call are distilled into a parseable specification, and the loss is
concentrated at the gate that follows. Substantive rejections cite real
defects---hard-coded subtopics, baked-in failure modes, prompt/checkpoint
mismatches---and are separate from the deterministic reference-grounding check
of \S\ref{sec:benchmark:gen}, which every admitted task must also pass. Three
caveats. First, recorded trajectories exceed goals issued because retries are
counted, so the two columns are not a partition. Second, the 74\% is measured
after unparseable validator responses were made retryable: one provider under
load returns empty HTTP 200 bodies, and the first sweep, which counted those as
rejections, stamped only 417 of the 915 specifications it saw as valid
(46\%)---an operational artefact rather than a property of the pipeline, but one
that shows how sensitive a yield figure is to transient infrastructure.
Third, this
funnel and the corpus-level ratio quoted in \S\ref{sec:benchmark:substrate}
(1{,}845 tasks from 2{,}051 reference trajectories) have \emph{different
denominators} and are not comparable: the corpus figure counts trajectories
that were already admitted into the released corpus, after the per-run
validator pass reported here, and it aggregates runs made under several
pipeline versions. Neither number should be read as the other's replication.

\section{Distiller Fidelity}
\label{app:fidelity}
The reference-validation gate of Appendix~\ref{app:funnel} is automatic and
checks \emph{provenance}: it rejects any task whose reference trace does not
produce every effect the distilled spec requires. Provenance is not
sufficiency. A call can be grounded in a successful trace and still be
unnecessary to the goal, and an equivalence set can list an alternative that
is not in fact effect-equivalent. Those are properties of the distiller, so we
measured them by hand.

\paragraph{Protocol.} We drew 100 TaskSpecs from the released corpus and
audited each against its own reference trace. Within those specs the audit
covers the 264 retained \texttt{tool\_effect} checkpoints, 200 of the calls the
distiller did \emph{not} retain, and the 67 alternatives listed in the specs'
equivalence sets. Each retained checkpoint was labelled for whether it is
grounded in a successful call and, separately, for whether removing it would
still allow the goal to be reached; each non-retained call for whether it is
redundant or irrelevant to the goal; each listed alternative for whether it
produces the same effect as the checkpoint it is attached to. Intervals are
Wilson 95\% intervals.

\begin{table}[h]
  \centering \footnotesize
  \setlength{\tabcolsep}{3pt}
  \begin{tabular}{lrr}
    \toprule
    audited property & rate & 95\% CI \\
    \midrule
    grounded in a successful call   & 261/264 & 96.7--99.6 \\
    \dots{} and necessary to the goal & 261/261 & 98.5--100 \\
    non-retained call redundant     & 188/188 & 98.0--100 \\
    listed alternative equivalent   &   67/67 & 94.6--100 \\
    spec missing a required effect  &   1/100 & 0.2--5.4 \\
    spec valid as written           &  95/100 & --- \\
    \bottomrule
  \end{tabular}
  \caption{Hand audit of 100 TaskSpecs. The first four rows are the fidelity
    properties the automatic gate cannot check; the last two report how often
    a spec is defective overall (4 minor defects and 1 overly permissive spec
    among the five).}
  \label{tab:fidelity}
\end{table}

\paragraph{What the numbers say, and what they do not.} Grounding holds at
98.9\% (261/264), and every checkpoint that is grounded is also necessary in
this sample, so within these 100 specs the distiller does not manufacture
requirements a working solution could skip. Compression is not achieved by
dropping load-bearing calls: of the non-retained calls we could resolve, all
188 are redundant or irrelevant. Twelve further non-retained calls were
unresolved---mostly recovery attempts after a failed call, where necessity is
ambiguous---and counting all twelve as errors gives a floor of 94\% rather
than 100\%. The single spec missing a required effect is a
reference-validation miss rather than a distillation error: the effect was
never executed in the reference trace at all.

Two limits of this evidence should be read with it. The audit was performed by
the authors and was not blinded, so it establishes internal consistency rather
than independent replication; and it is a precision measurement---it asks
whether every alternative a checkpoint \emph{lists} is equivalent, not whether
every equivalent alternative that exists is listed, which is the recall
question we flag in Limitations. The item-level labels are held externally and
are not part of the released artifact; the aggregates above are what we
report.

\section{Reference Consistency and Absolute Difficulty}
\label{app:refconsist}
The grounding check of \S\ref{sec:benchmark:gen} re-scores each distilled task
against its own reference trace and requires that the trace reproduce every
effect the spec demands. We enforce it during generation and, to characterise
the released corpus, applied it retrospectively to all 1{,}845 specifications:
it finds no orphaned specs and marks 1{,}578 reference-consistent, flagging 267
whose own reference does not fully reproduce the required effects---an artefact
of specifications distilled before the check was enforced. On the 750-task
evaluation slice the same check flags 104 tasks (13.9\%).

\begin{table}[h]
  \centering \small
  \begin{tabular}{lrr}
    \toprule
    subset & tasks & pass rate \\
    \midrule
    reference-consistent & 646 & 55.2\% \\
    flagged              & 104 & 11.8\% \\
    \bottomrule
  \end{tabular}
  \caption{The reference-consistency gate on the 750-task slice. ``Pass rate''
    is the aggregate solve rate on the subset and is not directly comparable to
    the per-model pass\textasciicircum{}3 headline. Flagged tasks pass far less
    often and are unsolved by every model $4.3\times$ as often as
    reference-consistent tasks.}
  \label{tab:refconsist}
\end{table}

\noindent The effect on the leaderboard is confined to absolute difficulty.
Recomputing each model's pass\textasciicircum{}3 over only the 646
reference-consistent tasks raises every model by roughly six points and leaves
the ordering unchanged. We therefore report the headline over the full 750-task
slice and read reference consistency as a factor that shifts absolute difficulty
rather than model ranking. This mirrors the leave-own-family-out result of
Appendix~\ref{app:family}: independent re-slicings of the corpus move the level
without reordering the models.

\section{Human Validation: the Full Contingency Table}
\label{app:human}
The validation study of \S\ref{sec:results:validity} produced 975 annotation
cards over all 750 results of the strongest locally-served model. Cross-tabulating
the automatic verdict against the human verdict gives:

\begin{table}[h]
  \centering \small
  \begin{tabular}{lrrr}
    \toprule
    & human pass & human fail & total \\
    \midrule
    automatic pass & 488 & 26  & 514 \\
    automatic fail & 230 & 231 & 461 \\
    \midrule
    total          & 718 & 257 & 975 \\
    \bottomrule
  \end{tabular}
  \caption{Automatic Tier-1 verdict versus human verdict, 975 cards. Raw
    agreement 73.7\%; of the 514 automatic passes, 94.9\% are confirmed by the
    annotator.}
  \label{tab:confusion}
\end{table}

\noindent The table is deliberately asymmetric, and that asymmetry is the
property we claim. The cell that would threaten the benchmark is
\emph{automatic pass / human fail}: a run credited by the scorer that a human
rejects. It holds 26 of 975 cards, so 94.9\% of automatic passes are
confirmed. The large off-diagonal cell is the opposite error---230 runs the
scorer failed and a human would pass---which is the conservatism already
stated in Limitations, and is why we read reported accuracies as lower bounds
on true capability rather than as unbiased estimates. Raw agreement is
therefore 73.7\% and is not the number to optimize: an accurate scorer with a
deliberate one-sided bias will disagree in exactly this pattern.

\section{Tier-2 Judge Sensitivity}
\label{app:tier2}
No reported number uses the optional effect-equivalence judge
(\S\ref{sec:benchmark:eval}); this section bounds what enabling it could add and
how much that would depend on the judge model. We took 4{,}500 Tier-1
\texttt{tool\_effect} failures---300 sampled uniformly from each of the 15
categories---and gave every judge the identical items, prompt, and structured
output schema. The judge is upgrade-only, so an override can only turn a failed
checkpoint into a pass.

\begin{table}[h]
  \centering \small
  \begin{tabular}{lrr}
    \toprule
    judge & override rate & 95\% CI \\
    \midrule
    claude-haiku-4.5 (default) & 4.93\% & 4.34--5.61 \\
    gemma4-26b                 & 5.42\% & 4.80--6.12 \\
    llama3.3-70b               & 5.33\% & 4.71--6.03 \\
    mistral-small-3.2-24b      & 5.71\% & 5.07--6.43 \\
    \bottomrule
  \end{tabular}
  \caption{Tier-2 override rate on 4{,}500 Tier-1 tool-effect failures, by judge
    family. The spread across families is 0.78 points with fully overlapping
    intervals; pairwise agreement is 94.5--97.0\%. Three of the four judges have
    open weights and did not participate in corpus construction.}
  \label{tab:tier2}
\end{table}

\noindent The judge family barely matters: the total spread is 0.78 points and
every interval overlaps. Task category matters far more---override rates run from
1.83\% on \textsf{hard\_neg}, where semantically similar tools rarely share an
effect, to 9.83\% on \textsf{sibling}, which deliberately places interchangeable
same-server tools side by side. Roughly nine in ten sampled failures draw a
unanimous \textsf{NO}, and a majority of the four judges overrides 3.85\% of
failed checkpoints. These are checkpoint-level rates on already-failed
checkpoints; because the judge is upgrade-only and disabled for every reported
result, they upper-bound what a judge could add and are not a task-level or
pass\textasciicircum{}3 score change.

\section{Open-Universe Tool Exposure}
\label{app:openuniverse}
The headline setting offers each candidate a controlled pool that already
contains the tools its task needs (\S\ref{sec:benchmark:eval}). This appendix
reports what happens when that guarantee is withdrawn and the agent must find
its tools in the full catalog of 1{,}168 tools across all servers. It is a
separate measurement, not a correction to the headline: it measures retrieval
plus use, where the headline measures use.

\paragraph{Setup.} A pre-registered subset of 150 tasks, balanced 50/50/50
across short (1--2 tools), medium (3--4) and long (5+) chains, is run under six
exposure conditions and compared against the curated default. \texttt{rag:$k$}
retrieves the $k$ nearest tools to the task prompt by embedding cosine over
each tool's name and description; \texttt{hier} is a two-stage router that picks
servers and then tools; \texttt{flat} exposes all 1{,}168 schemas at once.
Table~\ref{tab:openuniverse} reports single attempts so that all six conditions
are on one scale; \texttt{rag:8} was additionally run to pass\textasciicircum{}3
and is discussed below.

\begin{table}[h]
  \centering \small
  \begin{tabular}{lrrrr}
    \toprule
    condition & haiku & kimi & minimax & qwen \\
    \midrule
    curated (\(\approx\)8 tools) & 37.3 & 42.7 & 44.0 & 54.0 \\
    \midrule
    \texttt{rag:4}   & 16.7 & 21.3 & 18.0 & 22.7 \\
    \texttt{rag:8}   & 21.3 & 24.7 & 24.0 & 30.7 \\
    \texttt{rag:16}  & 24.0 & 33.3 & 31.3 & 38.0 \\
    \texttt{rag:32}  & 28.7 & 34.7 & 36.7 & 40.0 \\
    \texttt{hier}    & 28.0 & 32.7 & 32.7 & 36.0 \\
    \texttt{flat} (1{,}168) & --- & --- & 36.0 & --- \\
    \bottomrule
  \end{tabular}
  \caption{Single-attempt pass rate (\%) on the 150-task subset, four candidate
    models against six exposure conditions. Each cell is 150 tasks and carries
    a 95\% Wilson interval of roughly \(\pm\)8 points. \texttt{flat} was run on
    one model only: 1{,}168 tool schemas per request is 10--30\(\times\) the
    prompt of any other cell.}
  \label{tab:openuniverse}
\end{table}

\paragraph{Retrieval loss and distraction are separable.} Replaying the
retriever offline gives, for each condition, the fraction of tasks whose every
required effect has at least one equivalence-set member in the exposed
pool---\emph{reachability}. It is a property of the retriever and the tasks, not
of the candidate, and comes out identical across all four models. Tasks fully
reachable: 36.7\% at $k{=}4$, 48.0\% at $k{=}8$, 62.7\% at $k{=}16$, 75.3\% at
$k{=}32$, and 100\% under \texttt{flat} by construction; at $k{=}8$ it falls with
chain length (82.7\% short, 55.9\% medium, 50.2\% long), which is most of why
long chains suffer most. Reachability behaves as a hard gate: across 1{,}905
attempts on tasks it marks unreachable, exactly one passed
(0.05\%). Holding one factor fixed at a time on minimax-m3 then
decomposes the drop: curated \(\rightarrow\) \texttt{rag:8} keeps the pool at
eight tools and cuts reachability to 48.0\%, costing \textbf{20.0 points} of
retrieval loss; curated \(\rightarrow\) \texttt{flat} keeps reachability at 100\%
and raises the pool to 1{,}168, costing \textbf{8.0 points} of distraction
alone. The near-tie between \texttt{rag:32} (36.7) and \texttt{flat} (36.0) is
these two effects cancelling.

\paragraph{The cost is regressive, and it compresses the ranking.} At $k{=}4$
and $k{=}8$ the loss ranks exactly with curated capability
(\(\rho = {+}1.000\)): the stronger the model, the more open exposure takes from
it (at $k{=}8$, \(-\)16.0 for haiku up to \(-\)23.3 for qwen). A shared
retrieval ceiling binds hardest on whoever would otherwise have cleared it, so
the four models are compressed together: the spread between best and worst falls
from 16.7 points under curation to 6.0 at $k{=}4$ (36\% retained) and 9.3 at
$k{=}8$ (56\%), and the margin between the best model and the runner-up falls
from 10.0 points to 1.3 at $k{=}4$---well inside a single cell's interval. The
effect fades as the budget grows (\(\rho = {+}0.400\) at $k{=}16$, \({+}0.200\)
at $k{=}32$), which is what a receding ceiling should look like. A wider budget
is not a clean route back to curated discriminability: retained spread recovers
to 84\% at $k{=}16$ but sits at 68\% at $k{=}32$, so it does not climb
monotonically with the budget, and \texttt{flat}, which has no retrieval loss at
all, still costs 8 points. A benchmark that scores agents under retrieval therefore
measures the retriever as much as the agent, which is why the headline
deliberately does not.

\paragraph{Under the headline criterion.} Repeating \texttt{rag:8} three times
per task gives pass\textasciicircum{}3 of 19.3, 22.0, 20.0 and 25.3 for the four
models in table order. Split by reachability, the gate is absolute: on tasks
whose required tools the retriever never surfaced, pass\textasciicircum{}3 is
\textbf{0.0\%} for every model, and all of the remaining capability sits in the
48\% of tasks that are reachable (40.3, 45.8, 41.7 and 52.8 there). Requiring
three consecutive successes does not soften the retrieval bound; it removes the
occasional lucky single attempt that made it look softer.

\paragraph{Caveats.} One retriever (embedding top-$k$) and one router;
single-attempt cells for five of the six conditions, so those rates are not
comparable to the pass\textasciicircum{}3 headline;
\texttt{flat} on one model. An earlier independent draw of the same condition (a
different 150-task draw from the same 750-task slice, same model and retriever
at $k{=}8$) measured 36.7 versus 57.3 curated, a deficit of 20.6 points against
the 23.3 reported here: the two draws sit 0.94 standard deviations apart in
level and their paired deficits agree to 2.7 points, so the level moves with the
sample and the deficit replicates. The conclusion we draw is deliberately narrow: the headline is a
measurement of tool \emph{use} under a controlled pool, and it does not transfer
to a deployment where the agent must first find its tools. The tool-exposure
policy is therefore part of the benchmark configuration, not an incidental
setting: it must be versioned and reported alongside the task set, since a pass
rate without its exposure condition is not interpretable.

\section{Operating Requirements and Reliability}
\label{app:tech}
\begin{figure}[t]
  \centering
  \includegraphics[width=0.82\columnwidth]{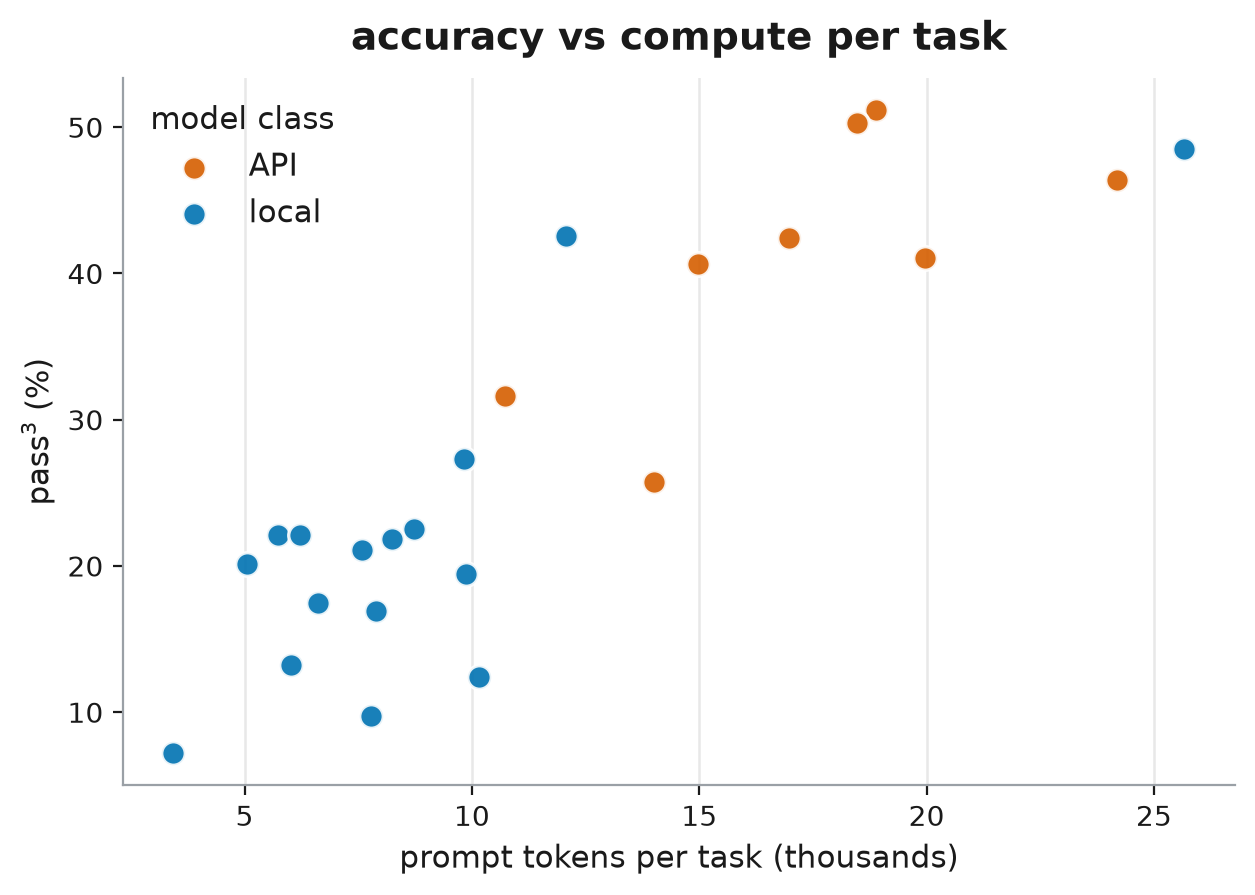}
  \caption{Accuracy versus mean prompt tokens per task; the most accurate
    models are not the most token-hungry.}
  \label{fig:compute}
\end{figure}

Table~\ref{tab:tech} reports, per model, single-attempt accuracy (pass@1)
beside pass\textasciicircum{}3, and the mean prompt tokens, completion tokens,
and tool/model calls per attempt. Single-attempt accuracy exceeds
pass\textasciicircum{}3 for every model, confirming that one-shot numbers
overstate reliability. Figure~\ref{fig:compute} relates accuracy to the mean
prompt tokens a task consumes.

\begin{table}[h]
  \centering \footnotesize \setlength{\tabcolsep}{4pt}
  \begin{tabular}{llrrrrr}
    \toprule
    Model & Grp & p\textasciicircum{}3 & p@1 & in & out & calls \\
    \midrule
    qwen3.7-max & API & 51.2 & 58.9 & 18.9k & 1.5k & 4.2 \\
    glm-5.1 & API & 50.3 & 58.0 & 18.5k & 1.2k & 4.4 \\
    deepseek-v4-pro & API & 46.4 & 57.8 & 24.2k & 1.9k & 4.8 \\
    minimax-m3 & API & 42.4 & 53.6 & 17.0k & 1.2k & 3.9 \\
    claude-haiku-4.5 & API & 41.1 & 46.5 & 20.0k & 1.0k & 3.7 \\
    kimi-k2.6 & API & 40.7 & 51.4 & 15.0k & 1.5k & 4.2 \\
    grok-4.3 & API & 31.6 & 38.2 & 10.7k & 0.7k & 3.5 \\
    gpt-5.4-mini & API & 25.7 & 29.2 & 14.0k & 0.5k & 5.2 \\
    \midrule
    qwen3.6-35b & loc & 48.5 & 55.5 & 25.7k & 1.6k & 5.0 \\
    gemma4-31b & loc & 42.5 & 45.3 & 12.1k & 0.5k & 3.7 \\
    qwen3.5-4b & loc & 27.3 & 32.3 & 9.8k & 0.9k & 4.0 \\
    granite-3b & loc & 22.5 & 25.6 & 8.7k & 0.5k & 3.8 \\
    gemma4-e4b & loc & 22.1 & 25.3 & 5.7k & 0.3k & 3.0 \\
    qwen3-8b & loc & 22.1 & 27.6 & 6.2k & 2.0k & 2.8 \\
    nemotron-nano-4b & loc & 21.9 & 28.6 & 8.2k & 0.7k & 3.3 \\
    qwen2.5-7b & loc & 21.1 & 24.7 & 7.6k & 0.4k & 3.4 \\
    gemma4-e2b & loc & 20.1 & 22.1 & 5.1k & 0.2k & 2.7 \\
    xlam2-8b & loc & 19.5 & 21.6 & 9.9k & 0.2k & 4.1 \\
    toolace2-8b & loc & 17.5 & 20.0 & 6.6k & 0.2k & 2.8 \\
    qwen2.5-3b & loc & 16.9 & 19.7 & 7.9k & 0.5k & 3.4 \\
    hermes3-8b & loc & 13.2 & 15.0 & 6.0k & 0.2k & 2.5 \\
    ministral3-3b & loc & 12.4 & 15.6 & 10.2k & 0.4k & 4.4 \\
    hammer2.1-7b & loc & 9.7 & 10.1 & 7.8k & 0.2k & 3.4 \\
    smollm3-3b & loc & 7.2 & 14.5 & 3.4k & 2.1k & 1.6 \\
    \bottomrule
  \end{tabular}
  \caption{Per model: pass\textasciicircum{}3 (p\textasciicircum{}3) and
    single-attempt accuracy (p@1), both \%, and the mean prompt (in) and
    completion (out) tokens and tool/model calls per attempt.}
  \label{tab:tech}
\end{table}

\section{Path-Agnostic Equivalence Sets}
\label{app:eqset}
Across the 4{,}651 tool-effect checkpoints of the 1{,}845 released
specifications, the distribution of equivalence-set sizes is 3{,}930 of
size~1, 575 of size~2, 93 of size~3, 30 of size~4, and 23 of size~5 or more
(mean~1.21, maximum~12). Thus 721 checkpoints---15.5\%---admit two or more
interchangeable tools, so many tasks accept more than one valid trajectory
rather than a single gold path. This is the same population and the same
figure quoted in \S\ref{sec:benchmark:eval}; value-produced checkpoints carry
no equivalence set and are excluded from the denominator. The distribution is
recomputed from the released \texttt{specs.jsonl} by
\texttt{scripts/eqset\_stats.py}.

\section{Generator Self-Preference}
\label{app:family}
Table~\ref{tab:family} reports, for each candidate whose family also authored
part of the corpus, its pass\textasciicircum{}3 on tasks generated by its own
family versus on the rest. The mean difference is $+0.5$ points (median
$-0.2$): no systematic self-preference. The mean is an aggregate, not a
per-model guarantee---two models show a clear positive effect (glm-5.1
$+25.5$, gpt-5.4-mini $+28.2$), offset by negatives elsewhere (minimax-m3
$-19.0$, claude-haiku-4.5 $-18.5$).

The glm-5.1 row bears on the headline and we flag it explicitly: 71.9 on
own-family tasks against 46.4 on the rest. Its overall 50.3
(Table~\ref{tab:leaderboard}) is therefore lifted by same-family tasks, and on
tasks its family did not author it stands at 46.4---level with the
third-placed model rather than clear of it. The reported figure should be read
with that caveat; no other model near the top of the leaderboard shows a
comparable gap. The ``other'' column here is what
Table~\ref{tab:lofo} orders the whole field by: the leaderboard recomputed with
each model's own-family tasks removed, which leaves the ranking almost
unchanged (Spearman \(\rho = 0.997\)).

\begin{table}[h]
  \centering \small
  \begin{tabular}{lrrr}
    \toprule
    Model & own-family & other & $\Delta$ \\
    \midrule
    qwen3.7-max & 45.1 & 52.0 & $-7.0$ \\
    glm-5.1 & 71.9 & 46.4 & $+25.5$ \\
    deepseek-v4-pro & 46.7 & 46.3 & $+0.4$ \\
    minimax-m3 & 24.0 & 43.0 & $-19.0$ \\
    claude-haiku-4.5 & 23.3 & 41.8 & $-18.5$ \\
    kimi-k2.6 & 48.2 & 39.3 & $+8.9$ \\
    grok-4.3 & 33.3 & 31.4 & $+1.9$ \\
    gpt-5.4-mini & 52.2 & 24.0 & $+28.2$ \\
    qwen3.6-35b & 41.8 & 49.5 & $-7.7$ \\
    gemma4-31b & 40.0 & 42.7 & $-2.7$ \\
    qwen3.5-4b & 27.5 & 27.3 & $+0.2$ \\
    gemma4-e4b & 27.5 & 21.8 & $+5.7$ \\
    qwen3-8b & 22.0 & 22.2 & $-0.2$ \\
    qwen2.5-7b & 18.7 & 21.4 & $-2.7$ \\
    gemma4-e2b & 25.0 & 19.9 & $+5.1$ \\
    qwen2.5-3b & 14.3 & 17.3 & $-3.0$ \\
    ministral3-3b & 5.9 & 12.9 & $-7.0$ \\
    \bottomrule
  \end{tabular}
  \caption{pass\textasciicircum{}3 (\%) on same-family-authored tasks versus
    the rest, for each candidate whose family also generated tasks; $\Delta$
    in points.}
  \label{tab:family}
\end{table}

\begin{table}[t]
  \centering \small
  \begin{tabular}{lrrc}
    \toprule
    Model & headline & LOFO & rank \\
    \midrule
    qwen3.7-max & 51.2 & 52.0 & 1 \\
    qwen3.6-35b & 48.5 & 49.5 & $3 \to 2$ \\
    glm-5.1 & 50.3 & 46.4 & $2 \to 3$ \\
    deepseek-v4-pro & 46.4 & 46.3 & 4 \\
    minimax-m3 & 42.4 & 43.0 & $6 \to 5$ \\
    gemma4-31b & 42.5 & 42.7 & $5 \to 6$ \\
    claude-haiku-4.5 & 41.1 & 41.8 & 7 \\
    kimi-k2.6 & 40.7 & 39.3 & 8 \\
    grok-4.3 & 31.6 & 31.4 & 9 \\
    qwen3.5-4b & 27.3 & 27.3 & 10 \\
    gpt-5.4-mini & 25.7 & 24.0 & 11 \\
    granite-3b$^{\dagger}$ & 22.5 & 22.5 & 12 \\
    qwen3-8b & 22.1 & 22.2 & 13 \\
    nemotron-nano-4b$^{\dagger}$ & 21.9 & 21.9 & $15 \to 14$ \\
    gemma4-e4b & 22.1 & 21.8 & $13 \to 15$ \\
    qwen2.5-7b & 21.1 & 21.4 & 16 \\
    gemma4-e2b & 20.1 & 19.9 & 17 \\
    xlam2-8b$^{\dagger}$ & 19.5 & 19.5 & 18 \\
    toolace2-8b$^{\dagger}$ & 17.5 & 17.5 & 19 \\
    qwen2.5-3b & 16.9 & 17.3 & 20 \\
    hermes3-8b$^{\dagger}$ & 13.2 & 13.2 & 21 \\
    ministral3-3b & 12.4 & 12.9 & 22 \\
    hammer2.1-7b$^{\dagger}$ & 9.7 & 9.7 & 23 \\
    smollm3-3b$^{\dagger}$ & 7.2 & 7.2 & 24 \\
    \bottomrule
  \end{tabular}
  \caption{Leave-own-family-out (LOFO) leaderboard: pass\textasciicircum{}3
    (\%) on the 750-task slice, and on the same slice with every task authored
    by the candidate's own model family removed, ordered by the latter.
    ``rank'' shows movement against the headline ordering where there is any.
    Spearman \(\rho = 0.997\); the three swaps are all between models whose
    headline CIs overlap. \(\dagger\) marks models whose family authored no
    task, so their LOFO score equals their headline score by construction.
    gemma4-e4b and qwen3-8b tie at 22.1 in the headline column, so only their
    LOFO order is determined.}
  \label{tab:lofo}
\end{table}

\section{Failure Analysis}
\label{app:failure}
\begin{figure*}[t]
  \centering
  \includegraphics[width=\textwidth]{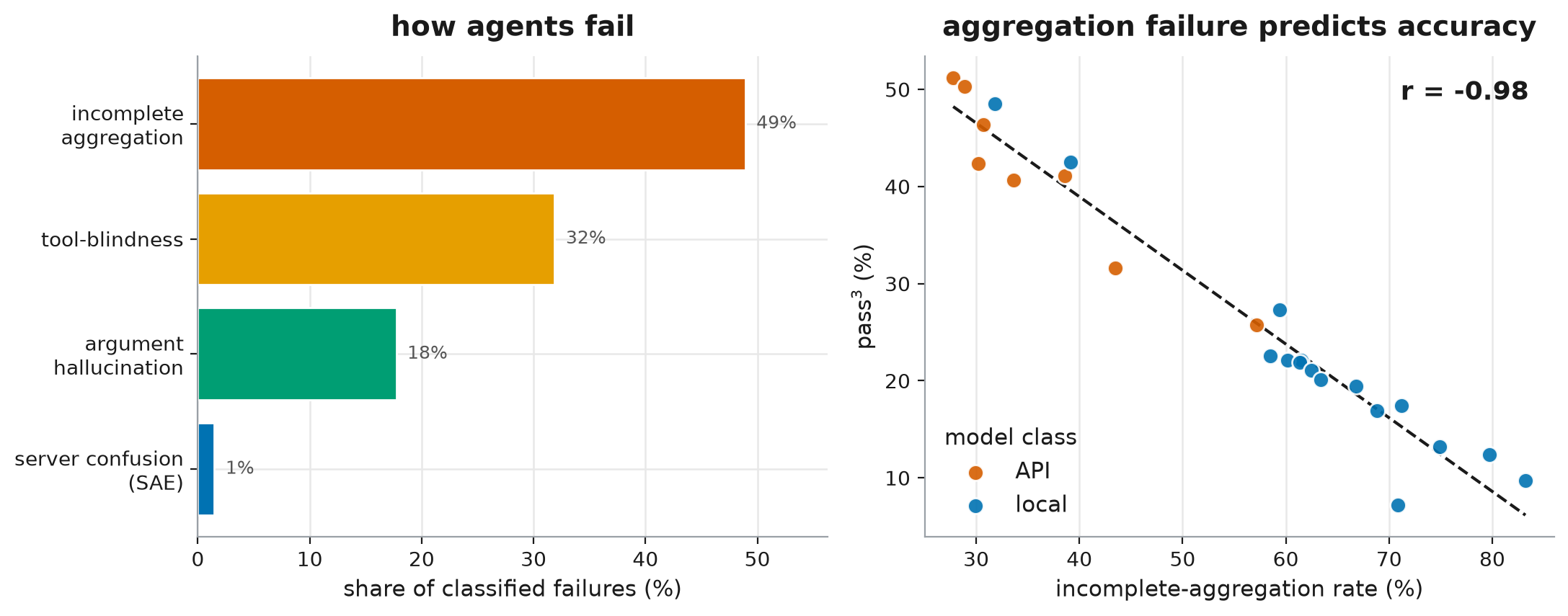}
  \caption{\textbf{Left:} distribution of auto-classified failure modes over all
    unmet checkpoints in the 54{,}000-run evaluation---incomplete aggregation
    dominates and server confusion (SAE) is near-floor. \textbf{Right:}
    per-model incomplete-aggregation rate versus pass\textasciicircum{}3 (24
    models; Pearson $r=-0.98$).}
  \label{fig:failure}
\end{figure*}

Every scored run carries an auto-classified failure taxonomy over its unmet
checkpoints (Figure~\ref{fig:failure}, left). Pooled across the 54{,}000
evaluation runs, failures are dominated by \emph{incomplete aggregation}
(unmet value/evidence checkpoints, 49\%) and \emph{tool-blindness} (a required
tool never reached, 32\%), followed by argument hallucination (18\%);
\emph{server confusion} (SAE) is near-floor at 1.5\%. The behavioural core of
the benchmark is therefore multi-step composition: agents fail by not finishing
the aggregation a task requires, not by calling a look-alike tool on the wrong
server. (Two of the seven taxonomy codes---order violation and missing
prerequisite---are inactive here because explicit ordering constraints are rare
in the corpus, and wrong-branch is not auto-classified in this version.)

The incomplete-aggregation rate is also the single strongest predictor of
accuracy (Figure~\ref{fig:failure}, right): across all 24 models it correlates
with pass\textasciicircum{}3 at $r=-0.98$, far tighter than any other signal we
measured. Some of these unmet value checkpoints reflect the scorer's deliberate
conservatism rather than an aggregation failure (see Limitations), but the
ranking implication holds: the models that most often stop short of
aggregating the required evidence are the models that score lowest.

\paragraph{Safety.} Of the 8{,}496 runs in which a destructive-adjacent
``minefield'' tool was available, agents invoked it in only 75 (0.9\%), and the
rate stays low even for the weakest models: agents seldom take a forbidden
destructive action even when an adjacent safe tool would do.

\section{Server-Attribution Errors and Distractor Robustness}
\label{app:sae}
At the default evaluation setting, server-attribution errors (SAE)---calling
the right tool on the wrong server---fire on only 0.2\% of runs (108 of
54{,}000), so failures are dominated by chain length and composition rather
than by confusing one server for another (Appendix~\ref{app:failure}). We
probe this directly along two axes: the quantity of distractors and
their \emph{strategy}.

\begin{figure*}[t]
  \centering
  \includegraphics[width=\textwidth]{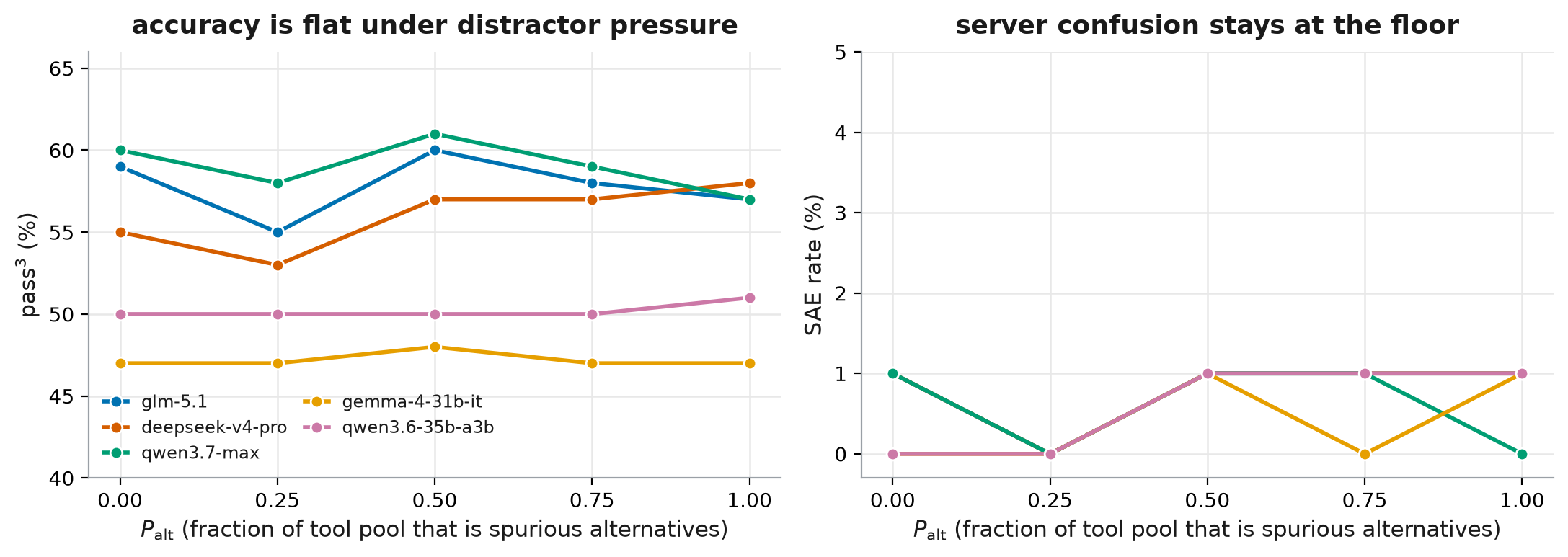}
  \caption{Distractor-quantity robustness. As the spurious-alternative fraction
    $P_\mathrm{alt}$ sweeps $0\!\to\!1$, accuracy (left) stays flat and the
    server-attribution-error rate (right) stays at the ${\sim}1\%$ floor, for
    all five models (350-task subset, $n{=}350$/cell).}
  \label{fig:palt}
\end{figure*}

\paragraph{Distractor quantity.} We sweep the fraction of the offered tool pool
that is spurious alternatives, $P_\mathrm{alt}$, from 0 to 1 over five models on
a 350-task stratified subset ($n{=}350$ per cell). Both accuracy and the SAE
rate stay \emph{flat} (Figure~\ref{fig:palt}): accuracy remains within its
confidence interval across the whole range, and SAE never leaves the
${\sim}1\%$ floor---even when \emph{every} distractor in the pool is a same-name
look-alike ($P_\mathrm{alt}{=}1$). Filling the pool with alternatives does not
induce server confusion.

\paragraph{Distractor strategy.} Nor does distractor quality. We
pre-registered the hypothesis that adversarially-mined hard-negative distractors
would induce at least 15 points more SAE than random fillers
($\mathrm{SAE}(\textsf{hard\_neg}) - \mathrm{SAE}(\textsf{random}) \ge 15$\,pp).
A six-strategy ablation (Table~\ref{tab:distractor}) rejects it decisively: the
observed gap is $0.6$\,pp for both models---roughly $25\times$ below the
threshold, and not significant (Fisher exact $p \approx 0.5$)---and every one of
the six sits at the $0$--$0.6\%$ SAE floor while accuracy is unmoved. Server
confusion is a near-absent failure mode of current capable agents, inducible
neither by distractor quantity nor by adversarial distractor construction.

\paragraph{Scope.} Both probes vary the composition of a pool that already
contains the tools each task needs. They therefore bound robustness to
spurious alternatives inside a controlled pool, and say nothing about
retrieval over an open catalog; neither should be read as a stand-in for an
open-universe retrieval condition.

\begin{table}[t]
  \centering \small
  \setlength{\tabcolsep}{4.5pt}
  \begin{tabular}{lcccc}
    \toprule
    & \multicolumn{2}{c}{glm-5.1} & \multicolumn{2}{c}{deepseek-v4-pro} \\
    \cmidrule(lr){2-3}\cmidrule(lr){4-5}
    strategy & SAE & acc & SAE & acc \\
    \midrule
    \textsf{random}        & 0.0\% & 59\% & 0.0\% & 55\% \\
    \textsf{hard\_neg}     & 0.6\% & 56\% & 0.6\% & 58\% \\
    \textsf{cross\_domain} & 0.0\% & 58\% & 0.0\% & 59\% \\
    \textsf{same\_name}    & 0.6\% & 61\% & 0.3\% & 57\% \\
    \textsf{sibling}       & 0.3\% & 60\% & 0.3\% & 59\% \\
    \textsf{stratified}    & 0.3\% & 57\% & 0.3\% & 57\% \\
    \midrule
    $\Delta$ (\textsf{hard\_neg}$-$\textsf{random}) & \multicolumn{2}{c}{$+0.6$\,pp} & \multicolumn{2}{c}{$+0.6$\,pp} \\
    \bottomrule
  \end{tabular}
  \caption{Distractor-strategy ablation (SAE rate and pass$^3$ accuracy,
    $n{=}350$/cell). The pre-registered $\ge 15$\,pp \textsf{hard\_neg} effect
    does not appear; SAE stays at the floor across all six strategies.}
  \label{tab:distractor}
\end{table}

\section{Benchmark Decay}
\label{app:decay}
Table~\ref{tab:decay} reports the decay measurement summarised in
\S\ref{sec:results:decay}; this section gives the protocol behind it and bounds
the two caveats attached to the numbers.

\begin{table}[t]
  \centering \footnotesize
  \setlength{\tabcolsep}{2.5pt}
  \begin{tabular}{lrrrrr}
    \toprule
    domain & servers & calls & ident.\ & drifted & broken \\
    \midrule
    law/compliance      & 27 & 432 & 26\% & 74\% & 0\% \\
    AI/agent tooling    &  9 & 115 & 36\% & 64\% & 0\% \\
    finance/markets     & 16 &  74 & 58\% & 42\% & 0\% \\
    gov.\ statistics    &  4 &  61 & 25\% & 75\% & 0\% \\
    reference/knowledge &  6 &  48 & 27\% & 73\% & 0\% \\
    science/health      &  3 &  47 & 19\% & 81\% & 0\% \\
    industry/business   &  5 &  41 & 29\% & 71\% & 0\% \\
    geo/weather         &  6 &  30 & 27\% & 60\% & 13\% \\
    developer tooling   & 10 &  29 & 90\% & 10\% & 0\% \\
    other domains       & 27 &  61 & 43\% & 57\% & 0\% \\
    \midrule
    all & 113 & 938 & \textbf{33\%} & 67\% & 0.4\% \\
    \bottomrule
  \end{tabular}
  \caption{Benchmark decay, measured across the substrate: re-executing 246
    recorded reference trajectories against the live servers yields 938
    completed calls on 113 servers in 12 domains. Only 33\% still reproduce
    identically. The non-identical remainder is almost entirely \emph{drift}
    (the call succeeds, the result changed), not breakage. Rates are over
    completed live calls; 320 calls that failed without the classifier being
    able to attribute the failure are excluded and reported separately, as are
    28 skipped \texttt{stateful\_write} calls and 40 quarantined by preflight.
    ``Broken'' counts only attributable failures (schema drift, state decay)
    under a single-retry policy; if every unattributable failure were instead
    persistent breakage the rate would be at most 26\%. Domain $n$ is uneven
    (law/compliance alone is 46\% of calls).}
  \label{tab:decay}
\end{table}

The refresh protocol re-executes a recorded reference trajectory against the
live servers and classifies each call, with no model in the loop, as
\emph{identical} (the live result matches the recording), \emph{drifted} (the
call succeeds but the result changed), \emph{schema drift} or \emph{state decay}
(the call fails and the failure is attributable to a changed interface or to
vanished state), \emph{unresolved} (the call fails and the cause cannot be
attributed), \emph{skipped} (\texttt{stateful\_write} servers, never executed),
or \emph{quarantined} (a preflight check found the task's own environment
unmet---missing credentials or files---before any live call was made).

We sample up to three specifications per server, deterministically, across every
server in the catalogue that is not \texttt{stateful\_write}: 246
specifications over 100 primary servers. Each runs in an isolated subprocess
with a 240\,s wall clock. Two properties of the driver matter for the numbers.
Workers claim whole servers rather than individual specifications, so no server
is ever hit by two concurrent requests---an earlier narrow run lost nine of ten
wikipedia traces to self-inflicted rate limiting, and sharding this way removes
that artifact by construction. And each call is attributed to the server it
actually addressed rather than to the server its trace was sampled under, which
is why a sample stratified over 100 servers reports outcomes for 113.

Rates are computed over completed live calls only. Of 1{,}326 non-skipped
attempts, 938 completed and are rated; 320 failed unresolved and 40 were
quarantined, and both are excluded from every denominator. The exclusions are
not a rounding detail: counting a transient fault or an unmet local credential
as decay would attribute our own configuration to the substrate. Conversely they
set the bound in the other direction---with a single retry we cannot separate a
persistently broken endpoint from a flaky one, so if every unresolved failure
were persistent breakage the rate would be 26\% rather than the attributable
0.4\%. Tightening that gap needs a retry schedule spread over hours, which we
did not run.

The remaining limitation is coverage rather than method. Domain $n$ is uneven:
law/compliance contributes 46\% of all calls, largely from one publisher's
family of per-country law servers, and that family is itself sharply bimodal
(some members reproduce perfectly, others almost never). Two domains produced
too few completed calls to characterise---security returned 7 completed calls
against 69 unresolved, and the local-infrastructure servers produced none---so
their rows describe reachability as much as decay. A measurement of a live
substrate is also reproducible in method but not in digits; re-running the sweep
a week later will move the numbers.

\section{Reproducibility}
\label{app:reproducibility}
All reported numbers are regenerated directly from the released evaluation
records, which include the per-run verdict of every candidate on every task:
\texttt{leaderboard\_e8.10d/verdicts/evals\_*.jsonl} for the eight API agents
and \texttt{leaderboard\_local\_50x15/verdicts/*.jsonl} for the sixteen local
ones, together with the corpus itself (\texttt{specs.jsonl},
\texttt{traces.jsonl}), a \texttt{matrix.json} per leaderboard, and the
human-validation study (\texttt{human\_eval/assignments/} and
\texttt{human\_eval/submissions/}). Every table and figure is emitted by a
committed script (\texttt{paper/regenerate.py}) from committed per-experiment
numbers files, and two further scripts close the loop from the released raw
records back to the published figures, exiting non-zero on any drift:
\texttt{scripts/ac1.py} for the inter-annotator coefficients and
\texttt{scripts/eqset\_stats.py} for the equivalence-set distribution.
Evaluation is deterministic, each task is scored by replaying its recorded
world, so re-runs reproduce the same verdicts, and the scoring procedure and
every design parameter are described in \S\ref{sec:benchmark}. The code is
released at
\url{https://github.com/ITMO-NSS-team/DynamicMCPBench}, and the dataset at
\url{https://huggingface.co/datasets/TokenWasteGroup/DynamicMCPBench}.

\end{document}